\pdfoutput=1

\documentclass[10pt,twocolumn,letterpaper]{article}

\usepackage[pagenumbers]{cvpr}

\definecolor{lpsOurs}{HTML}{D81B60}
\definecolor{lpsOursLight}{HTML}{F9D5E3}
\definecolor{lpsOursDark}{HTML}{A7144A}
\definecolor{lpsReference}{HTML}{7C8797}
\definecolor{lpsReferenceLight}{HTML}{B9C0CA}
\definecolor{lpsReferenceDark}{HTML}{4E5866}
\colorlet{lpsHighlight}{lpsOurs!12}

\DeclareRobustCommand{\lpsFigCheck}{\textcolor{lpsOursDark}{\ding{51}}}
\DeclareRobustCommand{\lpsFigCross}{\textcolor{lpsReferenceDark}{\ding{55}}}

\definecolor{rq1bg}{rgb}{1,0.85,0.85}
\definecolor{rq2bg}{rgb}{1,0.989548,0.874067}
\definecolor{rq3bg}{rgb}{0.85,1,0.85}
\definecolor{rq4bg}{rgb}{0.85,0.85,1}
\DeclareRobustCommand{\rqtag}[1]{RQ#1}
\DeclareRobustCommand{\rqhighlight}[2]{%
  \texorpdfstring{%
    \begingroup\sethlcolor{rq#1bg}%
    \hl{#2}%
    \endgroup}{#2}}
\DeclareRobustCommand{\rqheading}[2]{\rqhighlight{#1}{RQ#1: #2}}

\usepackage{microtype}

\usepackage{multirow}
\usepackage{longtable}
\usepackage{soul}
\usepackage{placeins}
\usepackage{colortbl}
\usepackage{pifont}
\DeclareRobustCommand{\cmark}{\textcolor{green!50!black}{\ding{51}}}
\DeclareRobustCommand{\xmark}{\textcolor{red!75!black}{\ding{55}}}
\usepackage{tikz}
\usepackage{pgfplots}
\pgfplotsset{compat=1.18}
\usetikzlibrary{shapes.geometric,arrows.meta,positioning,calc,backgrounds}
\usepackage{twemojis}
\usepackage[T1,OT1]{fontenc}
\usepackage{upquote}
\usepackage{tcolorbox}
\tcbuselibrary{breakable,skins,listings}
\newtcblisting{lpsprompt}[2][]{%
  enhanced,
  breakable,
  listing only,
  colback=lpsReference!5,
  colframe=lpsOursDark!60,
  colbacktitle=lpsOurs!10,
  coltitle=lpsOursDark,
  fonttitle=\sffamily\small\bfseries,
  boxrule=0.65pt,
  arc=3pt,
  left=6pt,
  right=6pt,
  top=4pt,
  bottom=4pt,
  toptitle=2.5pt,
  bottomtitle=2.5pt,
  before skip=5pt,
  after skip=5pt,
  title={#2},
  listing options={
    basicstyle=\fontencoding{T1}\ttfamily\scriptsize,
    columns=fullflexible,
    keepspaces=true,
    breaklines=true,
    breakatwhitespace=true,
    breakindent=8pt,
    showstringspaces=false,
    upquote=true,
    aboveskip=0pt,
    belowskip=0pt
  },
  #1
}
\newcommand{\paper}{Evolving Loop Programs}
\newcommand{\papershort}{ELP}
\newcommand{\papersubtitle}{Discovering How Frozen Transformers Should Loop}

\soulregister\paper{7}

\ifdefined\nolinenumbers\else
  \newenvironment{nolinenumbers}{}{}
\fi

\definecolor{cvprblue}{rgb}{0.21,0.49,0.74}
\usepackage[pagebackref,breaklinks,colorlinks,allcolors=cvprblue]{hyperref}

\title{\paper{}:\\
\papersubtitle{}}
\hypersetup{pdftitle={\paper{}: \papersubtitle{}}}

\definecolor{headerurl}{RGB}{208,25,66}
\newcommand{\PublicAuthorNames}{Yang Chen$^{1}$ \qquad Kaitai Shen$^{1}$ \qquad Xuemei Zhou$^{2}$ \qquad Teng Long$^{3}$ \qquad Vincent Tao Hu$^{1}$}
\newcommand{\PublicAffiliations}{{\normalsize $^{1}$~School of Software Engineering, Huazhong University of Science and Technology}\\[0.1em]
{\normalsize $^{2}$~Centrum voor Wiskunde en Informatica (CWI)} \qquad {\normalsize $^{3}$~University of Trento}}
\newcommand{\PublicProjectLink}{\\[0.55em]{\small Project Page: {\hypersetup{urlcolor=headerurl}\url{https://vibe-hust.github.io/elp_page/}}}\\[-0.05em]
{\small Code: {\hypersetup{urlcolor=headerurl}\url{https://github.com/vibe-hust/elp_code}}}}
\newcommand{\PublicAuthorBlock}{\PublicAuthorNames\\[0.35em]\PublicAffiliations\PublicProjectLink\vspace{-14pt}}
\newcommand{\PublicEmailFootnoteBox}{%
  \par\vspace{4pt}%
  \noindent\begin{minipage}{\columnwidth}
  \normalfont\footnotesize
  \rule{0.4\columnwidth}{0.4pt}\par\vspace{2pt}%
  \noindent Emails: Yang Chen (\href{mailto:chenyang4719@gmail.com}{\texttt{chenyang4719@gmail.com}}), Vincent Tao Hu (\href{mailto:taohu620@gmail.com}{\texttt{taohu620@gmail.com}}).
  \end{minipage}\par
}
\author{\PublicAuthorBlock}
\hypersetup{pdfauthor={Yang Chen, Kaitai Shen, Xuemei Zhou, Teng Long, Vincent Tao Hu},pdfsubject={Research preprint}}

\makeatletter
\g@addto@macro\@maketitle{\vspace{-14pt}
\begin{minipage}{\textwidth}
\normalfont\normalsize
\captionsetup{type=figure,skip=2pt}
\centering
\resizebox{0.93\textwidth}{!}{\input{figs/teaser_citations}}
\par\vspace{1pt}
\begin{minipage}{0.93\textwidth}
\begin{minipage}[b]{0.365\linewidth}
\centering
\includegraphics[width=\linewidth,trim=0 8pt 0 2pt,clip]{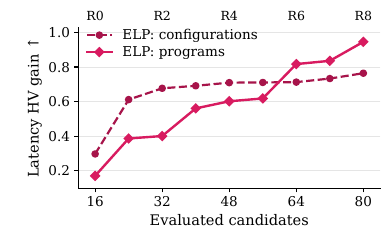}
\par\vspace{1pt}
{\small\bfseries (e) Both \papershort{} Variants Improve\par}
\end{minipage}\hfill
\begin{minipage}[b]{0.615\linewidth}
\centering
\includegraphics[width=\linewidth]{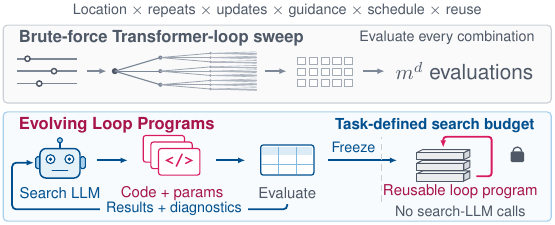}
\par\vspace{1pt}
{\small\bfseries (f) Searching Loop Programs\par}
\end{minipage}
\end{minipage}
\par\vspace{2pt}
\resizebox{0.93\textwidth}{!}{\input{figs/fig_teaser_longlive}}
\par\vspace{1pt}
{\small\bfseries (g) LongLive~2.0 Consistency\par}
\caption{\textbf{\paper{} (\papershort{}) improves image quality--compute trade-offs and long-video consistency without retraining.} Panels (a,e) use measured wall-clock speedup during Scale-RAE search; (b,c) use inference FLOPs for Looped-DiT. Search frontiers in (a,b) project three objectives. Selected programs deploy without the search LLM (c); frozen loop programs improve consistency across LongLive and Self-Forcing on RememBench~\cite{zhang2026mosaichunk} (d). Panel (e) reports cumulative latency-based hypervolume gain above the shared initial anchors, while (f) contrasts feedback-guided search with grid enumeration. Panel (g) shows a LongLive~2.0~(5B) video from departure ($3.6\,\mathrm{s}$) to revisit ($15.8\,\mathrm{s}$): the baseline collapses upon revisit, whereas our loop program restores the scene. Circle sizes in (c) are schematic; full-benchmark results include search examples, and external baselines in (c) follow~\cite{chng2026looped}.}
\label{fig:teaser}
\end{minipage}
\PublicEmailFootnoteBox
}
\makeatother

\input{reference-names}
\begin{document}
\addtocontents{toc}{\protect\setcounter{tocdepth}{-1}}
\hypersetup{hypertexnames=false}
\maketitle
\begin{abstract}
Reusing transformer blocks through loops can improve pretrained models without updating their weights. However, useful loops require coordinated choices about which blocks repeat, how hidden states evolve, and when computation is reused. We introduce \paper{} (\papershort{}), an offline LLM-based method for discovering executable inference programs with favorable quality--compute trade-offs. The search jointly edits loop structure, state-update dynamics, and execution controls, including sampling schedules, guidance, and intermediate-state reuse. The LLM writes diagnostic code and combines observations of model behavior with accumulated development evaluations to guide these edits. We retain nondominated programs for different operating priorities, then freeze their code and parameters. During deployment, these programs execute their loops without further calls to the search LLM. Experiments span image, video, and text using separately searched programs. For images, diffusion transformers trained with and without loops achieve better quality--compute trade-offs. A selected RAE program improves GenEval and DPG-Bench with a $5.19\times$ FLOPs speedup over standard unlooped inference. On Looped-DiT, both improve at equal or lower compute than native inference with matched sampling steps. For video, frozen programs improve long-video consistency across LongLive and Self-Forcing without weight updates. To the best of our knowledge, this is the first empirical validation of training-free transformer-block looping in pretrained video world models. For text, programs improve aggregate reasoning accuracy on Qwen3-4B. Our code will be released.
\end{abstract}
    
\section{Introduction}
\label{sec:intro}

Transformer research examines how depth and width affect computation and representations~\cite{levine2020limits}. Reusing blocks increases \emph{effective depth} without adding distinct parameters~\cite{dehghani2019universal}. Recurrent transformers have been trained for language modeling~\cite{geiping2025recurrent,bae2025mixture,li2026rectified} and image diffusion~\cite{chng2026looped,derakhshani2026lift}; training-free methods also introduce loops into frozen language models~\cite{chen2026trainingfreelooped}, flow-matching generators~\cite{yan2026trainingfree}, and dense-prediction backbones~\cite{chen2026gramloop}.

Useful loops, however, require coordinated choices. In autoregressive models, the sharing pattern, token-specific recursion depth, and key--value reuse affect how computation is allocated~\cite{bae2025mixture}. For frozen language models, damped state updates can succeed where naive block repetition degrades performance~\cite{chen2026trainingfreelooped}. In diffusion, the repeated layers, token scope, sampling-time gates, loop strength, and guidance~\cite{hu2023selfguided,krause2026guiding} all influence the resulting computation~\cite{yan2026trainingfree}. These choices also interact with the outer sampling schedule: more computation per denoiser call changes the budget for additional calls. Enumerating combinations requires repeated model evaluations, while fixed parameter menus exclude unimplemented update and reuse rules. We therefore seek useful \emph{loop programs}: executable rules and their parameters.

LLM-guided search offers a way to explore this space. Applications include robotic reward design~\cite{ma2024eureka}, combinatorial optimization~\cite{liu2024eoh,ye2024reevo}, and automated data science~\cite{guo2024dsagent}. Related efforts address mathematical discovery~\cite{romeraparedes2024funsearch}, prompt optimization~\cite{wang2024promptagent}, and optimizer adaptation~\cite{bai2026reactive}. Across these settings, researchers use evaluation feedback and task-specific search procedures to refine proposals, motivating a similar feedback loop for inference-program design. For loop discovery, a useful agent must reason about where repeated computation helps, how hidden states change, and which quality--cost compromises are acceptable. Aggregate scores assess outcomes; internal observations can suggest which program changes to test next.

We introduce \textbf{\paper{} (\papershort{})}. An LLM searches offline using development evaluations and diagnostics; selected code and settings are then frozen for deployment on new inputs without further search-LLM calls or program edits. Pretrained weights remain fixed throughout, so search reasoning and candidate-evaluation costs are incurred offline. The LLM jointly edits loop structure, state-update dynamics, and execution controls, including schedules, guidance, token selection, and intermediate-state reuse where supported. Four components organize the search: \emph{what to search}, executable loop programs; \emph{what to observe}, diagnostics written by the search LLM; \emph{how to search}, revisions informed by accumulated evaluations and observations; and \emph{what to evaluate and select}, task-specific objectives and nondominated trade-offs.

Figure~\ref{fig:teaser} separates search frontiers from final quality--compute and video comparisons. On RAE, a selected program improves GenEval and DPG with a $5.19\times$ FLOPs speedup over standard inference (Table~\ref{tab:main_results}). On Looped-DiT, both scores improve at equal or lower compute than native inference with matched sampling steps. Different tasks use separately searched programs, with multiple operating priorities within a search.

Our contributions are threefold:
\begin{itemize}
    \item We formulate training-free loop design as a search over executable programs, jointly covering loop structure, state-update dynamics, and execution controls.
    \item We develop a search workflow that combines programmable diagnostics, accumulated experimental feedback, and Pareto retention, and analyze how the search space, search LLM, search targets, and objectives affect observed trade-offs.
    \item Separately searched programs improve performance across \emph{image, video, and text}: image quality--compute trade-offs for diffusion transformers trained with and without loops, long-video consistency across LongLive~1.0, LongLive~2.0, and Self-Forcing, and aggregate Qwen3-4B reasoning accuracy. To the best of our knowledge, we provide the first empirical validation of training-free transformer-block looping in pretrained video world models, reducing mean LPIPS by $61.3\%$ on LongLive~2.0 across all $100$ RememBench scenes.
\end{itemize}

\section{Related Work}

\paragraph{Transformer looping.}
Transformer looping increases executed depth by reusing parameters, while training compute also depends on backpropagation depth~\cite{xiong2026truncate}. Universal Transformers learn a shared recurrent transformation~\cite{dehghani2019universal}, while Huginn scales latent reasoning through a recurrent core~\cite{geiping2025recurrent} and Ouro learns depth allocation during pretraining~\cite{zhu2025ouro}. Rectified Looped Transformer stabilizes recurrent language-model feedback by interpolating current states with predicted endpoints~\cite{li2026rectified}. For image generation, Looped-DiT trains repeated denoiser blocks with intermediate supervision and regulated attention~\cite{chng2026looped}, while LiFT supervises a shared DiT core with targets indexed by continuous computational depth~\cite{derakhshani2026lift}. For world modeling, LoopWM trains a shared transformer core with variable loop depth~\cite{lu2026loopedworld}. In contrast, we add transformer-block loops to frozen pretrained video world models without further training. Training-free methods extend repeated computation to frozen checkpoints. Training-Free Looped Transformers refine updates through damped residual substeps~\cite{chen2026trainingfreelooped}. Yan et al. control the layer range, token scope, sampling-time activation, and prediction-space guidance for flow-matching generators~\cite{yan2026trainingfree}; GramLoop gates replayed visual features using Gram consistency~\cite{chen2026gramloop}. Naive repetition can degrade performance~\cite{chen2026trainingfreelooped}, so useful loops require more than an increased repetition count. We search jointly for repetition, state-update, and sampling/reuse rules in both initially unlooped and pretrained looped backbones.

\paragraph{Knowledge discovery by agents.}
LLM agents have discovered mathematical constructions and bin-packing heuristics through FunSearch~\cite{romeraparedes2024funsearch}, and improved mathematical algorithms, data-center scheduling, and computation kernels through AlphaEvolve~\cite{novikov2025alphaevolve}. Such systems structure their search around executable representations and task-specific evaluation. FunSearch evolves a critical function within a program skeleton; EoH evolves natural-language design ideas together with code~\cite{liu2024eoh}, while ReEvo uses reflections on evaluated heuristics to guide later proposals~\cite{ye2024reevo}. OPAL extends this feedback loop to neural-network training: an agent rewrites both diagnostic and optimizer code during a run~\cite{bai2026reactive}. For multi-turn tool use, OODA-Tool separates task-state preservation from action generation through controller-checked intermediate states~\cite{guo2026oodatool}. This motivates preserving experimental evidence across successive program edits. Our LLM edits loop structure, dynamics, control, and diagnostic collectors between inference experiments with fixed weights. Task-defined evaluations guide nondominated retention; selected programs execute at deployment without the search LLM.

\section{Method}
\label{sec:method}

\subsection{Preliminaries and Problem Setting}
A transformer loop repeatedly applies the same block or block group $\mathbf{B}_{\boldsymbol{\theta}}$ to an evolving hidden state $\mathbf{h}$. The simplest loop is
\begin{equation}
\mathbf{h}^{(k+1)} = \mathbf{B}_{\boldsymbol{\theta}}\!\left(\mathbf{h}^{(k)}\right),
\quad k=0,\ldots,K-1.
\label{eq:basic_loop}
\end{equation}
Here, $\mathbf{h}^{(0)}$ is the input and $K$ is the number of passes; $K=1$ gives a single pass. In diffusion models, looping occurs within a denoiser call, before the sampler advances. Training-free approaches reuse pretrained blocks without updating their weights $\boldsymbol{\theta}$~\cite{yan2026trainingfree,chen2026gramloop}; Looped-DiT instead learns its weights with repeated blocks~\cite{chng2026looped}. Our search keeps weights fixed in both cases, either introducing loops into an unlooped model or reorganizing existing loops.

\begin{figure*}[t]
\centering
\resizebox{\textwidth}{!}{%
\input{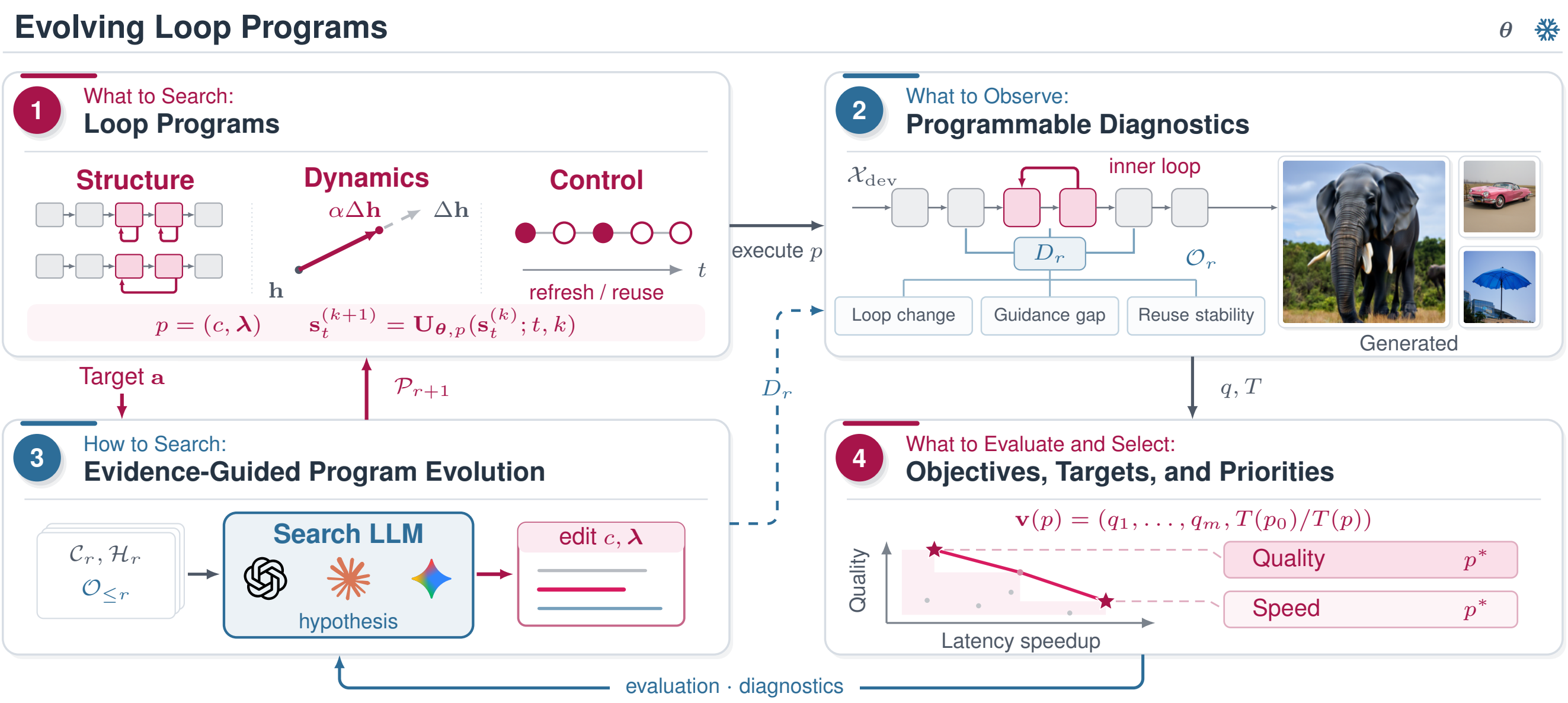}%
}
\caption{\textbf{\papershort{} evolves loop programs using diagnostics and development feedback.} Diagnostic evidence guides program edits; evaluation feeds back measured outcomes and supports Quality or Speed selection. Logos illustrate alternative search LLMs. Thumbnails show actual generated examples; the frontier is schematic. Pretrained weights remain fixed.}
\label{fig:method}
\end{figure*}

The LLM edits program code and settings offline using development evaluations and runtime diagnostics. Selected code and settings are frozen; deployment executes their loops on new inputs without further search-LLM calls. Figure~\ref{fig:method} shows the four components.

\subsection{\paper{}}

\paragraph{What to Search: Loop Programs.}
We search jointly over loop structure, state-update dynamics, and execution control, abbreviated as structure, dynamics, and control below. A candidate $p=(c,\boldsymbol{\lambda})$ contains executable code $c$ and its settings $\boldsymbol{\lambda}$. It generalizes the plain repetition in Eq.~\eqref{eq:basic_loop} to a programmable update:
\begin{equation}
\mathbf{s}^{(k+1)}_t
=\mathbf{U}_{\boldsymbol{\theta},p}\!\left(\mathbf{s}^{(k)}_t;t,k\right).
\label{eq:loop_program_update}
\end{equation}
Here, $t$ indexes inference steps, $k$ indexes repetitions, and $\mathbf{s}^{(k)}_t$ contains the hidden activations and any auxiliary state, such as cached residuals. The program determines which pretrained blocks $\mathbf{U}_{\boldsymbol{\theta},p}$ executes (structure), how their outputs update the state (dynamics), and when to repeat, stop, or reuse computation (control). The basic loop is recovered when the state contains only $\mathbf{h}$ and the update applies $\mathbf{B}_{\boldsymbol{\theta}}$ once. These decisions interact with sampling schedules, guidance, token selection, and intermediate-state reuse, subject to the interfaces supported by each backbone. Editing both code and settings allows the search LLM to introduce loop mechanisms beyond a predefined configuration space. Pretrained weights, task interfaces, and the benchmark scoring protocol remain fixed.

\paragraph{What to Observe: Programmable Diagnostics.}
The search LLM can write diagnostic collectors to inspect candidate behavior through intermediate states and execution traces on development inputs. At round $r$, $D_r$ produces
\begin{equation}
\mathcal{O}_r(p)
=D_r\!\left(\operatorname{Trace}_{\boldsymbol{\theta}}(p,\mathcal{X}_{\mathrm{dev}})\right),
\label{eq:program_diagnostics}
\end{equation}
where $\operatorname{Trace}_{\boldsymbol{\theta}}$ exposes runtime states from executing $p$ on development inputs $\mathcal{X}_{\mathrm{dev}}$, and $\mathcal{O}_r(p)$ contains collected tensors or summaries. Observations include changes across repetitions, differences between guidance branches, and the temporal stability of reusable states. The LLM may revise $D_r$ as new hypotheses arise. For example, stable residuals across neighboring sampling steps may motivate a program that periodically refreshes and otherwise reuses them. These observations guide proposed edits, whose usefulness is assessed by evaluating the resulting programs. Collectors must preserve the behavior being measured.

\paragraph{How to Search: Evidence-Guided Program Evolution.}
At each search round, the LLM uses accumulated evidence to formulate a testable hypothesis and edit program code and settings. Its available context includes source code, instructions, and conversation history $\mathcal{C}_r$; past evaluations and failures $\mathcal{H}_r$; and collected diagnostics $\mathcal{O}_{\leq r}$. We summarize the multi-turn proposal of candidate batch $\mathcal{P}_{r+1}$ as
\begin{equation}
\mathcal{P}_{r+1}
=\operatorname{LLM}\!\left(\mathcal{C}_r,\mathcal{H}_r,\mathcal{O}_{\leq r};\mathbf{a}\right),
\label{eq:program_evolution}
\end{equation}
where the target vector $\mathbf{a}$, defined below, guides the proposals. The LLM can coordinate edits across several design dimensions. Before execution, proposals are validated and duplicate code--configuration pairs are rejected. Each evaluated version is fixed so that its outcome can be associated with a specific program and configuration. Measurements and failures are added to the search history for subsequent reasoning. The outer search loop edits the program offline; the inner transformer loops run during candidate evaluation and later deployment.

\paragraph{What to Evaluate and Select: Objectives, Targets, and Priorities.}
Different applications favor speed, quality, or a compromise between them. We implement control over these \textbf{priorities} through objective configuration, target prompting, and representative selection. First, we configure which measured objectives enter the coordinator's ranking and Pareto feedback. For joint quality--speed search, we use
\begin{equation}
\mathbf{v}(p)=\left(q_1(p),\ldots,q_m(p),\frac{T(p_0)}{T(p)}\right),
\label{eq:search_objectives}
\end{equation}
where the quality scores $q_j(p)$ are oriented so that larger is better, $T(p)$ is measured inference latency, and $p_0$ is the fixed baseline. Single-objective image searches rank candidates only by the chosen score. In quality-only text search, the coordinator excludes latency from both ranking and LLM feedback. FLOPs-based efficiency is computed separately for reporting.

Second, the LLM receives a \textbf{target vector} $\mathbf{a}$ specifying the objective values it is encouraged to pursue. We vary \textbf{target strength} by raising these coordinates relative to a measured reference; the image-search experiments report their mean relative increase. We retain the full vector because equal average strength can emphasize quality or speed differently. These targets steer proposals without changing measured outcomes or the fixed hypervolume reference point.

During search, the coordinator returns measured scores and the Pareto frontier, or the best candidates under the configured single objective. A program is nondominated if no evaluated alternative is at least as good on every objective and strictly better on one; hypervolume (HV) summarizes the resulting trade-offs under fixed objectives and a fixed reference point. This feedback lets the LLM compare the consequences of its edits under the configured objectives.

Finally, the joint image-search prompt explicitly requests \textbf{Speed}, \textbf{Balanced}, and \textbf{Quality} representatives with supporting evidence. The LLM recommends evaluated programs favoring efficiency, a compromise, or task quality, respectively. One joint search can therefore supply several operating points. The text-reasoning comparison reports one frozen representative; alternative programs and their selection criteria are detailed in the supplement. Objective inclusion and ranking are enforced by the coordinator; priority-specific recommendations do not impose universal scalar weights or hard latency bounds. Supplementary Section~\ref{sec:supp_search_design} records the recommendations and replay identities. We freeze each selected program's code and settings for confirmation and deployment without further calls to the search LLM.

\section{Experiments}
\label{sec:exp}

We evaluate image generation (\rqtag{1}), search design and discovered programs (\rqtag{2}), video generation (\rqtag{3}), and text reasoning (\rqtag{4}), then report offline search cost separately from deployment cost.

\begin{tcolorbox}[enhanced,colback=lpsReference!3,colframe=lpsReference!45,
    boxrule=0.45pt,arc=3pt,boxsep=0pt,
    left=3.5pt,right=3.5pt,top=3pt,bottom=3pt,
    before skip=4pt,after skip=3pt,
    title={TL;DR},fonttitle=\bfseries,coltitle=lpsOursDark,
    attach boxed title to top left={xshift=6pt,yshift*=-\tcboxedtitleheight/2},
    boxed title style={colback=lpsOursLight!35,colframe=lpsReference!45,
        boxrule=0.45pt,arc=2pt,boxsep=0pt,
        left=4pt,right=4pt,top=1pt,bottom=1pt}]
\begin{itemize}[leftmargin=*,nosep,topsep=1pt,itemsep=0.5pt]
    \item \textbf{\rqhighlight{1}{Better images with less computation.}} On RAE, a selected program improves both compositional image-quality scores while using approximately \textbf{one-fifth of the baseline FLOPs}.
    \item \textbf{\rqhighlight{1}{Models already trained with loops still benefit.}} On Looped-DiT, changing how the loops execute improves both image-quality scores \textbf{at unchanged compute}.
    \item \textbf{\rqhighlight{3}{Long videos revisit scenes more consistently.}} On LongLive~2.0, perceptual mismatch between departure and revisit frames decreases by \textbf{61.3\% across 100 scenes}, although this improvement requires additional computation.
    \item \textbf{\rqhighlight{4}{The benefits extend to language reasoning.}} On the evaluated \textbf{4,145-question suite}, Qwen3-4B accuracy increases from \textbf{63.81\% to 65.48\%}.
\end{itemize}
\noindent\emph{Pretrained weights stay frozen. Task-specific programs run without further search-LLM calls.}
\end{tcolorbox}

\subsection{Experimental Setup}
\label{sec:exp_setup}

\paragraph{Models and tasks.}
We evaluate Scale-RAE~\cite{tong2026scaling} and the DiffusionBench RAE-SigLIP2-B/DDT 615M adaptation~\cite{leng2026diffusionbench}, both trained without loops, and Looped-DiT-B32~\cite{chng2026looped}, trained with a four-pass weight-tied core. RAE here is distinct from the official RAEv2 release. We also test autoregressive video models LongLive 2.0 5B~\cite{chen2026longlive2}, LongLive 1.0 1.3B~\cite{yang2026longlive}, and Self-Forcing 1.3B~\cite{huang2025selfforcing}, short-video diffusion model Wan2.1-1.3B~\cite{wan2025wan}, and text model Qwen3-4B~\cite{yang2025qwen3}. All pretrained weights remain frozen. We reran standard inference for these backbones, including the native Looped-DiT settings, under the corresponding evaluation protocols. We also reran Train-Free Loop without and with LoopGuidance~\cite{yan2026trainingfree} on Scale-RAE and RAE, and Fixed Loop~\cite{chen2026trainingfreelooped} on Qwen3-4B.

\paragraph{Benchmarks and metrics.}
We measure image composition with GenEval~\cite{ghosh2023geneval} and DPG-Bench~\cite{hu2024ella}; long-video consistency with CLIP, LPIPS, and TempSSIM on $100$ RememBench T2V scenes~\cite{zhang2026mosaichunk}; and short-video quality with VBench~\cite{huang2024vbench} on a local $256$-video collection. Text evaluation covers $4{,}145$ questions from local MMLU-STEM~\cite{hendrycks2021measuring}, GPQA-Main~\cite{rein2024gpqa}, ARC-Challenge~\cite{clark2018think}, and SciQ~\cite{welbl2017crowdsourcing}; Micro Acc weights questions equally, whereas Macro Group weights benchmarks equally. Text and short-video findings concern these evaluated instances. FLOPs speedup reports baseline-to-program denoiser operation ratios, excluding offline search cost; measured wall-clock speedups corroborate these ratios in Supplementary Sections~\ref{sec:supp_benchmark_results} and~\ref{sec:supp_looped_dit}. Development hypervolume uses GenEval, DPG, and measured latency speedup for Scale-RAE, or FLOPs speedup for Looped-DiT; it is not compared across backbones.

\paragraph{Search protocol and data splits.}
Image searches evaluate $80$ candidates over nine rounds. Scale-RAE target studies and Looped-DiT-B32 use $320$ development and $1{,}298$ withheld prompts; Full combines both and is not independent confirmation. LongLive uses $20$ search and $80$ held-out scenes. Speed and Quality denote representative priorities, not unique Pareto optima. Selected programs are frozen for replay without search-LLM calls. Detailed protocols and hardware are provided in Supplementary Section~\ref{sec:supp_exp_settings}.

\subsection{Results and Analysis}

Table~\ref{tab:main_comparison} provides a unified summary of our frozen-program comparisons across image generation (a,b), text reasoning (c), and long-video consistency (d).

\begin{table*}[t]
\centering
\caption{\textbf{\papershort{} improves image quality--compute trade-offs and long-video and text quality.} In (b), Looped-DiT~\cite{chng2026looped} uses its original inference rule ($K{=}4$ loop passes per sampling step). Video cells report mean (median); long-video consistency gains use additional compute. Scale-RAE stock-LG costs use reconstructed call counts (Supplementary Section~\ref{sec:supp_unlooped_results}). Gray shading: loop programs; gray text: unlooped baselines and the unranked Ultra variant. \textbf{Bold}/\underline{underlined}: best/second-best within each network. $\mathrm{ELP}_{\mathrm{quality}}^{\mathrm{ultra}}$: Ultra-target Quality from Table~\ref{tab:rae_synthetic_target_ablation}, unranked.}
\label{tab:main_comparison}
\begin{subtable}[t]{0.488\textwidth}
\centering
\caption{\textbf{\rqhighlight{1}{Text-to-Image Diffusion: Models Trained without Loops.}}}
\label{tab:main_results}
\renewcommand{\arraystretch}{1.12}
\footnotesize
\setlength{\tabcolsep}{2pt}
\begin{tabular}{@{}l r | r c c | c c@{}}
\toprule
\multirow{2}{*}{\textbf{Method}} & \multirow{2}{*}{\textbf{\#step}} & \multirow{2}{*}{\textbf{TFLOPs $\downarrow$}} & \multicolumn{2}{c|}{\textbf{Speedup}} & \textbf{GenEval} & \textbf{DPG} \\
 & & & \textbf{Base} & \textbf{w/ LG} & \textbf{(\%) $\uparrow$} & \textbf{(\%) $\uparrow$} \\
\midrule
\multicolumn{7}{@{}l}{\textit{Backbone: Scale-RAE}~\cite{tong2026scaling}} \\
\textcolor{black!55}{Baseline} & \textcolor{black!55}{25} & \textcolor{black!55}{31.58} & \textcolor{black!55}{$1.00\times$} & \textcolor{black!55}{$2.77\times$} & \textcolor{black!55}{51.64} & \textcolor{black!55}{77.21} \\
Loop w/o LG~\cite{yan2026trainingfree} & 25 & 56.04 & $0.56\times$ & $1.56\times$ & 57.04 & \underline{80.00} \\
Loop + LG~\cite{yan2026trainingfree} & 25 & 87.62 & $0.36\times$ & $1.00\times$ & 58.56 & 79.71 \\
\midrule
\rowcolor{gray!12}
$\mathrm{ELP}_{\mathrm{speed}}$ (Ours) & 10 & \textbf{19.56} & $\boldsymbol{1.61\times}$ & $\boldsymbol{4.48\times}$ & \underline{60.54} & \textbf{80.19} \\
\rowcolor{gray!12}
$\mathrm{ELP}_{\mathrm{quality}}$ (Ours) & 14 & \underline{27.13} & \underline{$1.16\times$} & \underline{$3.23\times$} & \textbf{60.98} & \textbf{80.19} \\
\textcolor{black!55}{$\mathrm{ELP}_{\mathrm{quality}}^{\mathrm{ultra}}$ (Ours)} & \textcolor{black!55}{16} & \textcolor{black!55}{31.23} & \textcolor{black!55}{$1.01\times$} & \textcolor{black!55}{$2.81\times$} & \textcolor{black!55}{61.63} & \textcolor{black!55}{80.22} \\
\midrule
\multicolumn{7}{@{}l}{\textit{Backbone: RAE}~\cite{leng2026diffusionbench}} \\
\textcolor{black!55}{Baseline} & \textcolor{black!55}{50} & \textcolor{black!55}{29.07} & \textcolor{black!55}{$1.00\times$} & \textcolor{black!55}{$1.86\times$} & \textcolor{black!55}{38.56} & \textcolor{black!55}{70.91} \\
Loop w/o LG~\cite{yan2026trainingfree} & 50 & 39.67 & $0.73\times$ & $1.37\times$ & 38.87 & 72.44 \\
Loop + LG~\cite{yan2026trainingfree} & 50 & 54.20 & $0.54\times$ & $1.00\times$ & 43.28 & \underline{74.68} \\
\midrule
\rowcolor{gray!12}
$\mathrm{ELP}_{\mathrm{speed}}$ (Ours) & 6 & \textbf{5.60} & $\boldsymbol{5.19\times}$ & $\boldsymbol{9.68\times}$ & \underline{44.42} & \textbf{74.96} \\
\rowcolor{gray!12}
$\mathrm{ELP}_{\mathrm{quality}}$ (Ours) & 10 & \underline{9.33} & \underline{$3.12\times$} & \underline{$5.81\times$} & \textbf{44.94} & 74.47 \\
\bottomrule
\end{tabular}
\end{subtable}%
\hfill
\begin{subtable}[t]{0.488\textwidth}
\centering
\setcounter{subtable}{2}%
\caption{\textbf{\rqhighlight{4}{Text Reasoning: Qwen3-4B.}}}
\label{tab:llm_reasoning}
\renewcommand{\arraystretch}{1.12}
\small
\setlength{\tabcolsep}{3pt}
\begin{tabular*}{\linewidth}{@{\extracolsep{\fill}}l | c c >{\columncolor{gray!12}[\tabcolsep][0pt]}c@{}}
\toprule
\textbf{Metric (\%) $\uparrow$} & \shortstack{\textbf{Non-}\\\textbf{looped}~\cite{yang2025qwen3}} & \shortstack{\textbf{Fixed}\\\textbf{Loop}~\cite{chen2026trainingfreelooped}} & \shortstack{\textbf{ELP}\\\textbf{(Ours)}} \\
\midrule
\multicolumn{4}{@{}l}{\textit{Individual benchmarks}} \\
MMLU-STEM Acc & 62.43 & \underline{64.20} & \textbf{64.72} \\
GPQA-Main Acc & 31.92 & \underline{33.26} & \textbf{34.82} \\
ARC Acc & \underline{50.60} & \textbf{51.71} & \textbf{51.71} \\
ARC Acc Norm & 54.18 & \textbf{54.95} & \underline{54.61} \\
SciQ Acc & 95.70 & \underline{96.20} & \textbf{96.50} \\
\midrule
\multicolumn{4}{@{}l}{\textit{Aggregate scores}} \\
Micro Acc & 63.81 & \underline{65.04} & \textbf{65.48} \\
Micro Acc Norm & 64.27 & \underline{65.72} & \textbf{65.91} \\
Macro Group & 60.16 & \underline{61.34} & \textbf{61.94} \\
Macro Domain & 58.37 & \underline{59.60} & \textbf{60.36} \\
\bottomrule
\end{tabular*}
\end{subtable}
\par\medskip
\begin{subtable}[t]{0.488\textwidth}
\centering
\setcounter{subtable}{1}%
\caption{\textbf{\rqhighlight{1}{Text-to-Image Diffusion: Models Trained with Loops.}}}
\label{tab:looped_dit_main}
\renewcommand{\arraystretch}{1.15}
\footnotesize
\setlength{\tabcolsep}{1.35pt}
\begin{tabular}{@{}l l r | r c | c c@{}}
\toprule
\multirow{2}{*}{\textbf{Setting}} & \multirow{2}{*}{\textbf{Method}} & \multirow{2}{*}{\textbf{\#step}} & \textbf{TFLOPs} & \textbf{Speedup $\uparrow$} & \textbf{GenEval} & \textbf{DPG} \\
 & & & \textbf{$\downarrow$} & \textbf{vs.\ Full} & \textbf{(\%) $\uparrow$} & \textbf{(\%) $\uparrow$} \\
\midrule
\begin{tabular}[c]{@{}l@{}}Full\\reference\end{tabular} & Looped-DiT & 100 & 53.18 & $1.00\times$ & \underline{84.65} & \textbf{85.34} \\
\midrule
 & Looped-DiT & 14 & \underline{7.45} & \underline{$7.14\times$} & 79.36 & 81.31 \\
\rowcolor{gray!12}
\cellcolor{white}\multirow{-2}{*}{\shortstack[l]{Matched\\14-step}} & \mbox{$\mathrm{ELP}_{\mathrm{speed}}$ (Ours)} & 14 & \textbf{6.80} & $\boldsymbol{7.82\times}$ & 83.43 & \underline{85.11} \\
\midrule
 & Looped-DiT & 25 & 13.30 & $4.00\times$ & 83.16 & 84.76 \\
\rowcolor{gray!12}
\cellcolor{white}\multirow{-2}{*}{\shortstack[l]{Matched\\25-step}} & \mbox{$\mathrm{ELP}_{\mathrm{quality}}$ (Ours)} & 25 & 13.30 & $4.00\times$ & \textbf{85.08} & 85.09 \\
\bottomrule
\end{tabular}
\end{subtable}%
\hfill
\begin{subtable}[t]{0.488\textwidth}
\centering
\setcounter{subtable}{3}%
\caption{\textbf{\rqhighlight{3}{Long Video Generation on RememBench T2V.}}}
\label{tab:world_model}
\label{tab:remembench_longlive20_5b}
\label{tab:remembench_longlive10_13b}
\label{tab:remembench_self_forcing_13b}
\begingroup
\scriptsize
\setlength{\tabcolsep}{1.1pt}
\renewcommand{\arraystretch}{1.35}
\begin{tabular*}{\linewidth}{@{\extracolsep{\fill}}l | c c c@{}}
\toprule
\textbf{Method} & \textbf{CLIP $\uparrow$} & \textbf{LPIPS $\downarrow$} & \textbf{TempSSIM $\uparrow$} \\
 & \textbf{Mean (Median)} & \textbf{Mean (Median)} & \textbf{Mean (Median)} \\
\midrule
LongLive 2.0 5B~\cite{chen2026longlive2} & \underline{0.8585} (\underline{0.9018}) & \underline{0.3852} (\underline{0.3910}) & \underline{0.9681} (\underline{0.9718}) \\
\rowcolor{gray!12}
\quad ELP (Ours) & \textbf{0.9282} (\textbf{0.9568}) & \textbf{0.1492} (\textbf{0.0974}) & \textbf{0.9905} (\textbf{0.9940}) \\
\midrule
LongLive 1.0 1.3B~\cite{yang2026longlive} & \underline{0.8670} (\underline{0.8919}) & \underline{0.2609} (\underline{0.2591}) & \underline{0.9819} (\underline{0.9839}) \\
\rowcolor{gray!12}
\quad ELP (Ours) & \textbf{0.9218} (\textbf{0.9526}) & \textbf{0.1480} (\textbf{0.1363}) & \textbf{0.9869} (\textbf{0.9903}) \\
\midrule
Self-Forcing 1.3B~\cite{huang2025selfforcing} & \underline{0.6585} (\underline{0.6705}) & \underline{0.4874} (\underline{0.4915}) & \underline{0.9780} (\underline{0.9801}) \\
\rowcolor{gray!12}
\quad ELP (Ours) & \textbf{0.7617} (\textbf{0.7764}) & \textbf{0.3607} (\textbf{0.3427}) & \textbf{0.9804} (\textbf{0.9827}) \\
\bottomrule
\end{tabular*}
\endgroup
\end{subtable}
\end{table*}

\paragraph{\rqheading{1}{Can \papershort{} Improve Image Generation?}}
\label{sec:exp_main}

For image generation, compare panels (a) and (b) of Table~\ref{tab:main_comparison}: the former introduces loops into models trained without them, while the latter reorganizes computation in a model already trained with repeated blocks.

\paragraph{Models Trained without Loops.}
Prior training-free looping~\cite{yan2026trainingfree} improves composition over standard inference but slows generation dramatically because it layers extra block passes and full reference calls onto an unchanged outer schedule. Table~\ref{tab:main_results} shows that \papershort{} breaks this accuracy--latency trade-off by reallocating compute across denoising time: compositional layout crystallizes in high-noise early steps while late steps refine local texture, so concentrating recurrence early while reusing guidance corrections or conditioning streams lets $10$-step Scale-RAE and $6$-step RAE programs surpass full-step Loop~+~LG at $4.48\times$ and $9.68\times$ fewer FLOPs ($1.61\times$ and $5.19\times$ faster than standard inference). Quality programs allocate additional steps to push GenEval further, though with diminishing DPG returns on RAE.

\begin{figure*}
\centering
\begingroup
\setlength{\tabcolsep}{1pt}
\renewcommand{\arraystretch}{1.0}
\ifdefined\qualwidth\else\newlength{\qualwidth}\fi
\setlength{\qualwidth}{0.0933\textwidth}
\newcommand{\qualgood}[1]{\textcolor{lpsOursDark}{\textbf{#1}}}
\newcommand{\qualbad}[1]{\textcolor{lpsReferenceDark}{#1}}
\newcommand{\qualdetail}[7]{%
  \begin{minipage}[t]{\dimexpr\qualwidth+1pt\relax}\centering
  \begin{tikzpicture}[x=\qualwidth,y=\qualwidth]
    \node[anchor=south west,inner sep=0pt] at (0,0)
      {#1};
    \draw[white,line width=0.8pt] ({#3/224},{1-#6/224}) rectangle ({#5/224},{1-#4/224});
    \draw[#7,line width=0.35pt] ({#3/224},{1-#6/224}) rectangle ({#5/224},{1-#4/224});
    \draw[#7,line width=0.45pt] (0,0) rectangle (1,1);
  \end{tikzpicture}\par\vspace{1.5pt}
  \pgfmathsetlengthmacro{\qscaledwidth}{224/(#5-#3)*\qualwidth}
  \pgfmathsetlengthmacro{\qcropheight}{(#6-#4)/(#5-#3)*\qualwidth}
  \pgfmathsetlengthmacro{\qxshift}{-#3/(#5-#3)*\qualwidth}
  \pgfmathsetlengthmacro{\qyshift}{-(224-#6)/(#5-#3)*\qualwidth}
  \begin{tikzpicture}
    \path[use as bounding box] (0,0) rectangle (\qualwidth,\qcropheight);
    \begin{scope}
      \clip (0,0) rectangle (\qualwidth,\qcropheight);
      \node[anchor=south west,inner sep=0pt] at (\qxshift,\qyshift)
        {#2};
    \end{scope}
    \draw[#7,line width=0.45pt] (0,0) rectangle (\qualwidth,\qcropheight);
  \end{tikzpicture}
  \end{minipage}%
}

\newcommand{\qualcomparisonheader}{%
\multicolumn{5}{c}{\footnotesize\textbf{(a) Scale-RAE}} &
\multicolumn{5}{c}{\footnotesize\textbf{(b) RAE}} \\[3pt]
\scriptsize 25-step Base & \scriptsize 10-step Base & \scriptsize Loop w/o LG & \scriptsize Loop + LG & \scriptsize \textbf{Ours (10-step)} &
\scriptsize 50-step Base & \scriptsize 10-step Base & \scriptsize Loop w/o LG & \scriptsize Loop + LG & \scriptsize \textbf{Ours (10-step)} \\[1pt]
\footnotesize 31.58 & \footnotesize 12.63 & \footnotesize 56.04 & \footnotesize 87.62 & \footnotesize\qualgood{19.56} &
\footnotesize 29.07 & \footnotesize 5.81 & \footnotesize 39.67 & \footnotesize 54.20 & \footnotesize\qualgood{9.33} \\[1pt]
\multicolumn{5}{c}{\scriptsize TFLOPs $\downarrow$} & \multicolumn{5}{c}{\scriptsize TFLOPs $\downarrow$} \\[4pt]
}

\scriptsize
\newcommand{\qualmethodcost}[2]{\shortstack[c]{#1\\[1pt]{\footnotesize #2}}}
\begin{tabular}{@{}ccccc@{\hspace{7pt}}ccccc@{}}
\multicolumn{5}{@{}p{0.49\textwidth}@{}}{\footnotesize\textbf{(a) Scale-RAE: Text rendering}\hfill\scriptsize\textcolor{lpsReferenceDark}{TFLOPs $\downarrow$}} &
\multicolumn{5}{@{}p{0.49\textwidth}@{}}{\footnotesize\textbf{(b) RAE: Color attribution}\hfill\scriptsize\textcolor{lpsReferenceDark}{TFLOPs $\downarrow$}} \\[1pt]
\multicolumn{5}{@{}l}{\emph{``\ldots vibrant red heavy truck \ldots stop sign \ldots''}} &
\multicolumn{5}{@{}l@{}}{\emph{``a photo of a purple wine glass and a black apple''}} \\[3pt]
\qualmethodcost{25-step Base}{31.58} &
\qualmethodcost{10-step Base}{12.63} &
\qualmethodcost{Loop w/o LG}{56.04} &
\qualmethodcost{Loop + LG}{87.62} &
\qualmethodcost{\textbf{Ours}}{\qualgood{19.56}} &
\qualmethodcost{50-step Base}{29.07} &
\qualmethodcost{10-step Base}{5.81} &
\qualmethodcost{Loop w/o LG}{39.67} &
\qualmethodcost{Loop + LG}{54.20} &
\qualmethodcost{\textbf{Ours (10-step)}}{\qualgood{9.33}} \\[3pt]
\qualdetail{\includegraphics[width=\qualwidth]{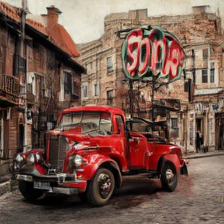}}{\includegraphics[width=\qscaledwidth]{figs/qualitative/sr_dpg_372_base_full.jpg}}{96}{12}{208}{100}{lpsReferenceDark} &
\qualdetail{\includegraphics[width=\qualwidth]{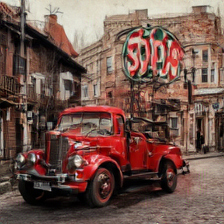}}{\includegraphics[width=\qscaledwidth]{figs/qualitative/sr_dpg_372_base_s10.jpg}}{96}{12}{208}{100}{lpsReferenceDark} &
\qualdetail{\includegraphics[width=\qualwidth]{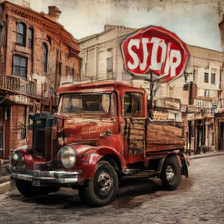}}{\includegraphics[width=\qscaledwidth]{figs/qualitative/sr_dpg_372_loop_nolg.jpg}}{96}{12}{208}{100}{lpsReferenceDark} &
\qualdetail{\includegraphics[width=\qualwidth]{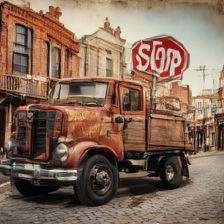}}{\includegraphics[width=\qscaledwidth]{figs/qualitative/sr_dpg_372_loop_lg.jpg}}{96}{12}{208}{100}{lpsReferenceDark} &
\qualdetail{\includegraphics[width=\qualwidth]{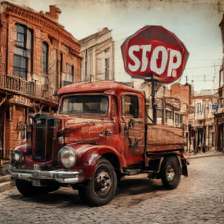}}{\includegraphics[width=\qscaledwidth]{figs/qualitative/sr_dpg_372_ours.jpg}}{96}{12}{208}{100}{lpsOurs} &
\qualdetail{\includegraphics[width=\qualwidth]{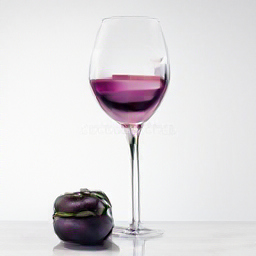}}{\includegraphics[width=\qscaledwidth]{figs/qualitative/rv_ge_453_base_full.jpg}}{0}{92}{168}{224}{lpsReferenceDark} &
\qualdetail{\includegraphics[width=\qualwidth]{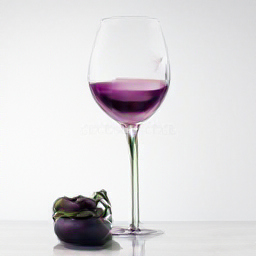}}{\includegraphics[width=\qscaledwidth]{figs/qualitative/rv_ge_453_base_s10.jpg}}{0}{92}{168}{224}{lpsReferenceDark} &
\qualdetail{\includegraphics[width=\qualwidth]{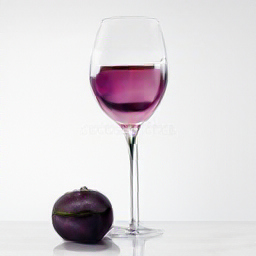}}{\includegraphics[width=\qscaledwidth]{figs/qualitative/rv_ge_453_loop_nolg.jpg}}{0}{92}{168}{224}{lpsReferenceDark} &
\qualdetail{\includegraphics[width=\qualwidth]{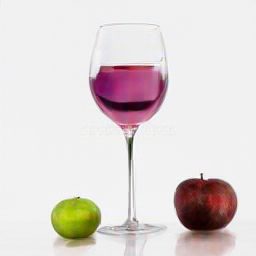}}{\includegraphics[width=\qscaledwidth]{figs/qualitative/rv_ge_453_loop_lg.jpg}}{42}{92}{210}{224}{lpsReferenceDark} &
\qualdetail{\includegraphics[width=\qualwidth]{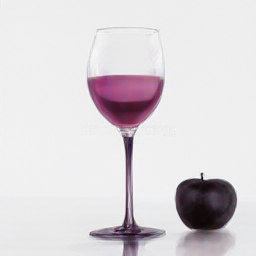}}{\includegraphics[width=\qscaledwidth]{figs/qualitative/rv_ge_453_ours.jpg}}{56}{92}{224}{224}{lpsOurs} \\
\noalign{\vskip 4pt}
\scriptsize\qualbad{``SOOR'' (0.80)} & \scriptsize\qualbad{Garbled (0.80)} & \scriptsize\qualbad{``SJOR'' (0.90)} & \scriptsize\qualbad{``SLOP'' (0.80)} & \scriptsize\qualgood{``STOP'' (1.00)} &
\scriptsize\qualbad{\lpsFigCross{} (Blob)} & \scriptsize\qualbad{\lpsFigCross{} (Blob)} & \scriptsize\qualbad{\lpsFigCross{} (Plum)} & \scriptsize\qualbad{\lpsFigCross{} (2 apples)} & \scriptsize\qualgood{\lpsFigCheck{} (Black apple)} \\
\end{tabular}
\endgroup
\par
\vspace{4pt}
\begingroup
\setlength{\tabcolsep}{1.5pt}
\newcommand{\lpsvideoframe}[2]{%
  \begin{tikzpicture}[baseline=(frame.center)]
    \node[inner sep=0pt] (frame)
      {#2};
    \draw[#1,line width=.4pt] (frame.south west) rectangle (frame.north east);
  \end{tikzpicture}%
}
\newcommand{\lpsvideolabel}[3]{%
  \raisebox{-.5\height}{\footnotesize\textcolor{#1}{\shortstack[l]{#2\\[-1pt]#3}}}%
}
\noindent\makebox[\textwidth][l]{%
  \footnotesize\textbf{(c) LongLive 2.0 (5B): Video frames}%
  \hfill\scriptsize\textcolor{lpsReferenceDark}{Video time (s)}%
}\par\vspace{1pt}
\begin{tabular*}{\textwidth}{@{\extracolsep{\fill}}lcccccc@{}}
& \footnotesize 0s & \footnotesize 3s & \footnotesize 6s & \footnotesize 9s & \footnotesize 12s & \footnotesize 15s \\[1pt]
\lpsvideolabel{lpsReferenceDark}{Baseline}{(no loop)} &
\lpsvideoframe{lpsReferenceDark}{\includegraphics[width=.148\textwidth]{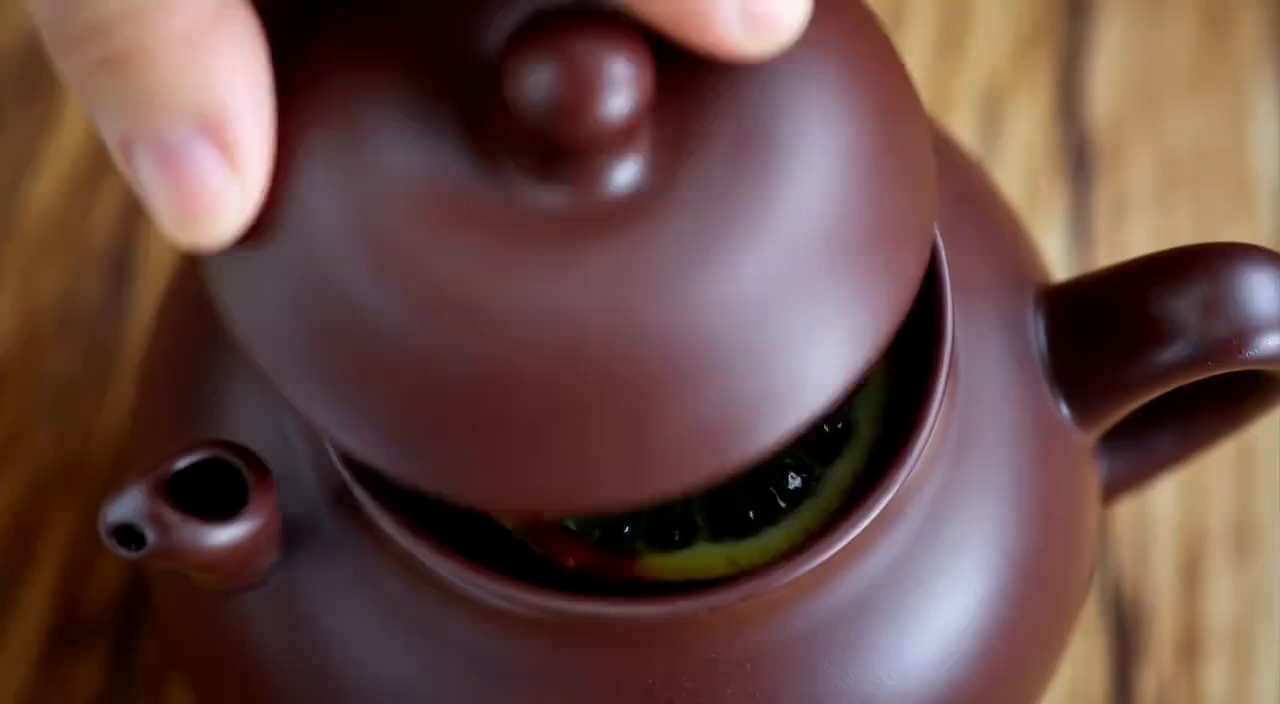}} &
\lpsvideoframe{lpsReferenceDark}{\includegraphics[width=.148\textwidth]{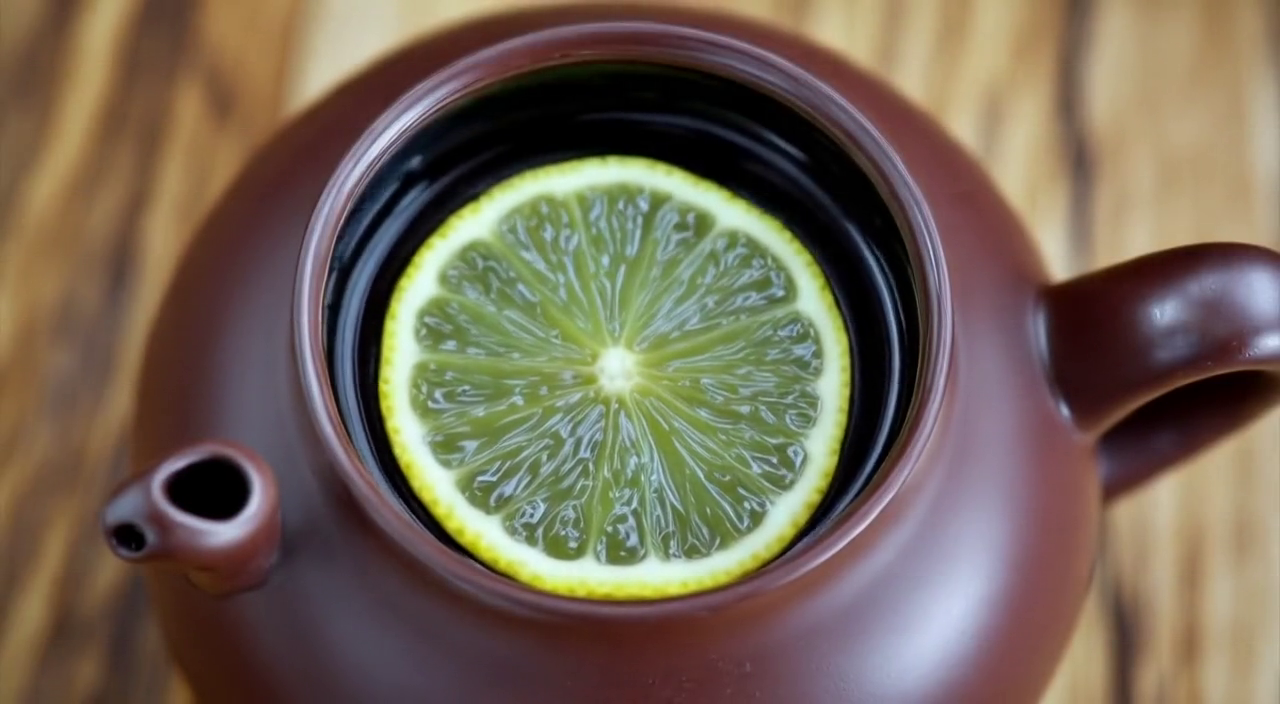}} &
\lpsvideoframe{lpsReferenceDark}{\includegraphics[width=.148\textwidth]{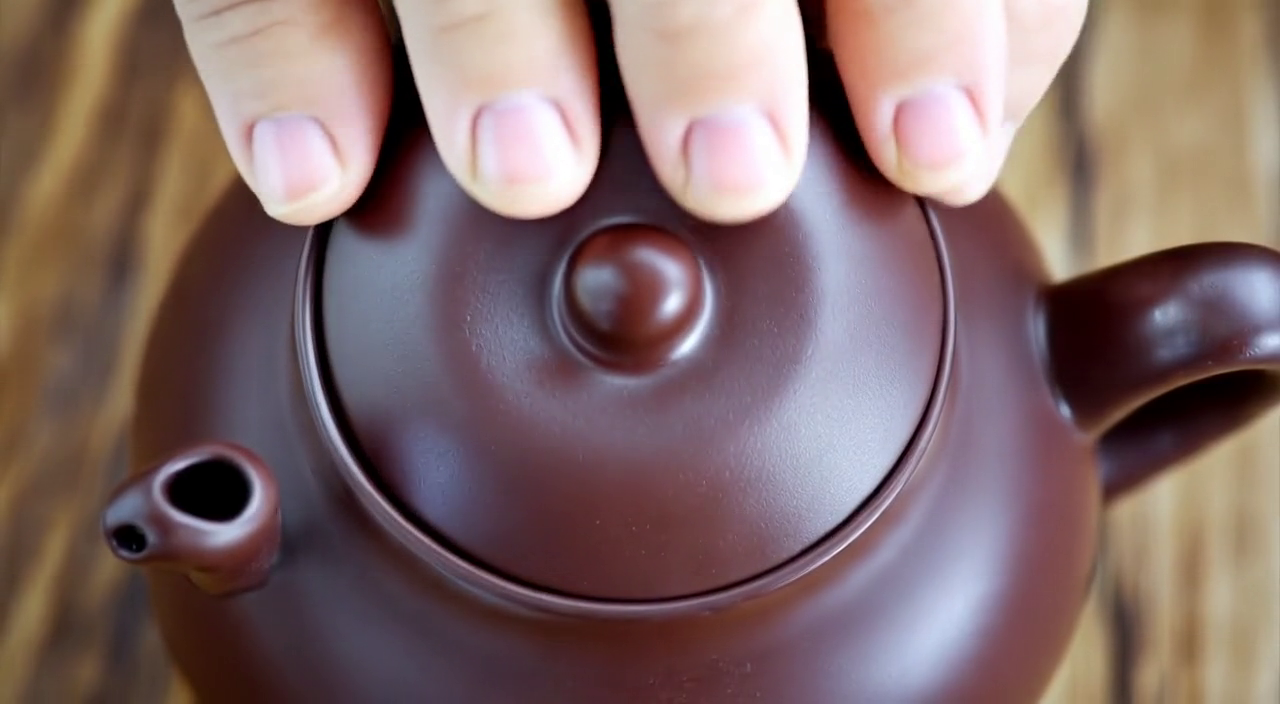}} &
\lpsvideoframe{lpsReferenceDark}{\includegraphics[width=.148\textwidth]{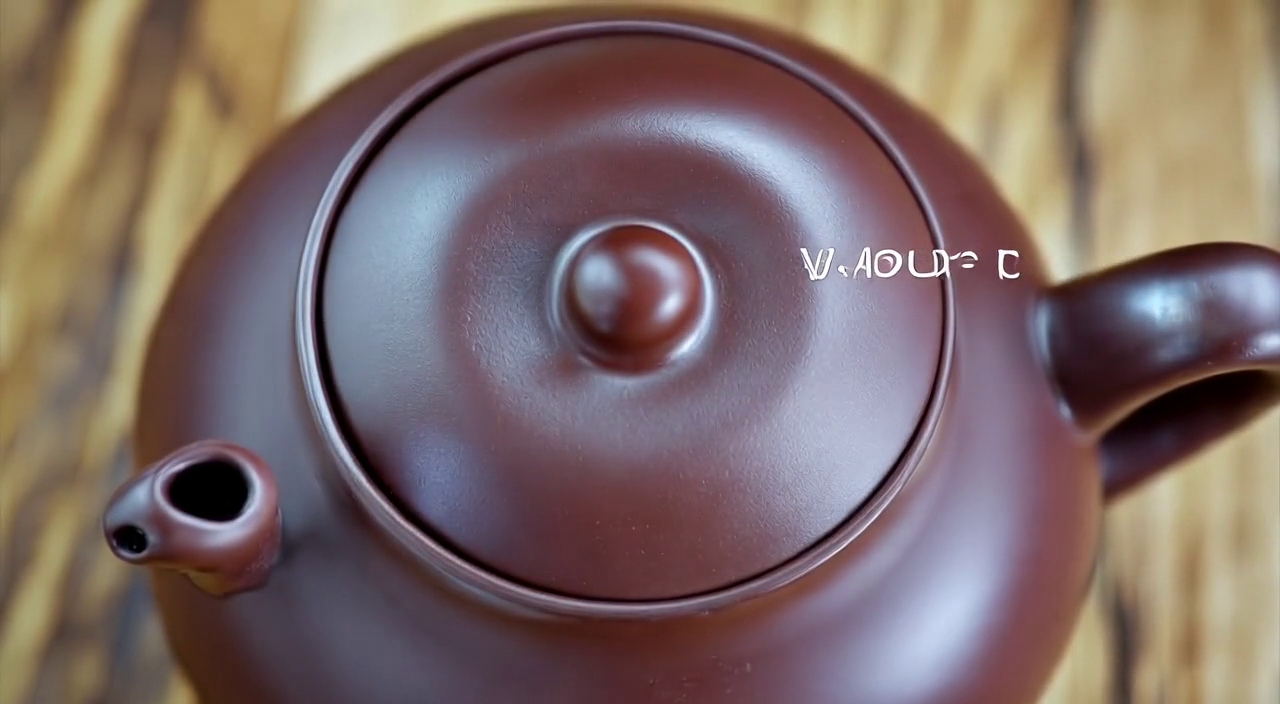}} &
\lpsvideoframe{lpsReferenceDark}{\includegraphics[width=.148\textwidth]{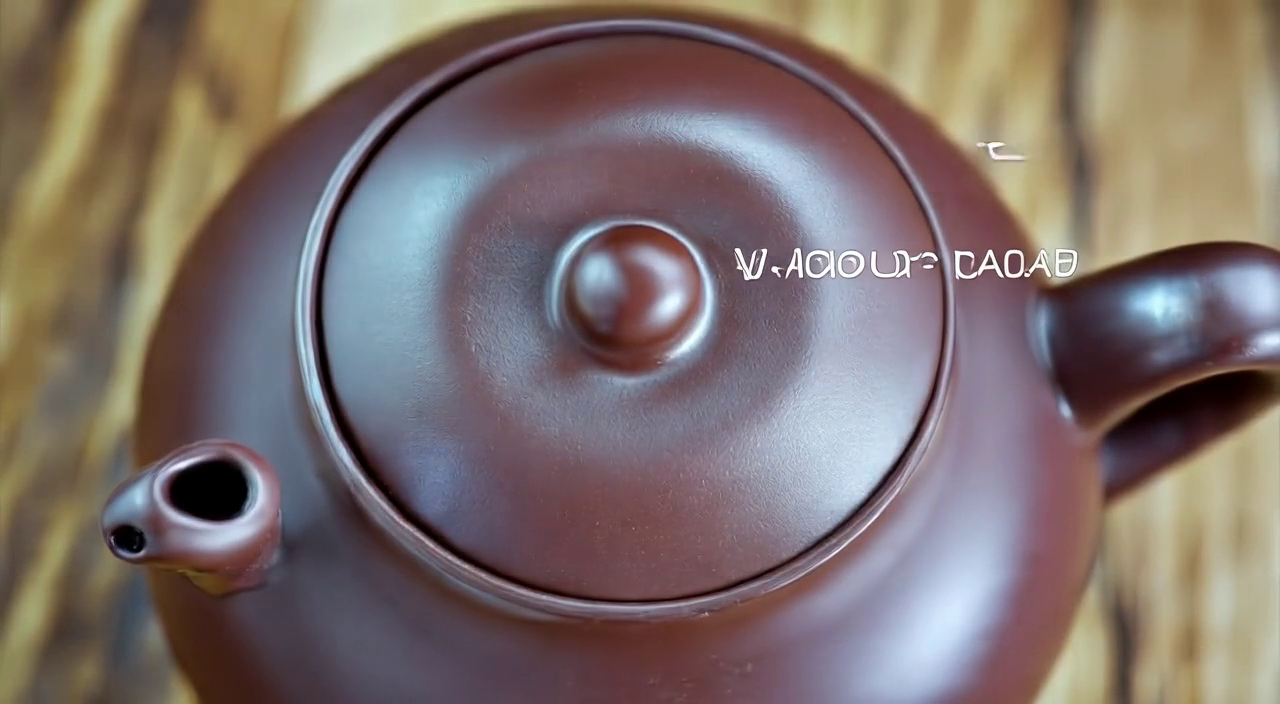}} &
\lpsvideoframe{lpsReferenceDark}{\includegraphics[width=.148\textwidth]{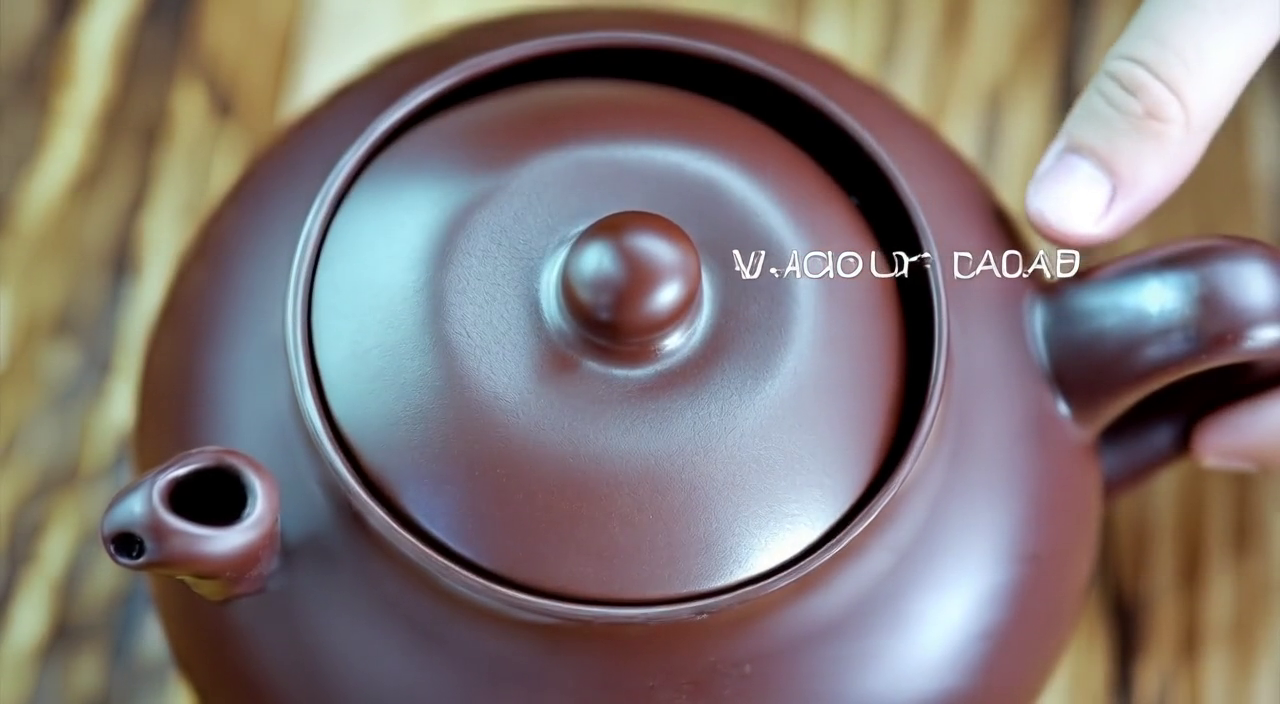}} \\
\noalign{\vskip2pt}
\lpsvideolabel{lpsOursDark}{ELP}{(Ours)} &
\lpsvideoframe{lpsOurs}{\includegraphics[width=.148\textwidth]{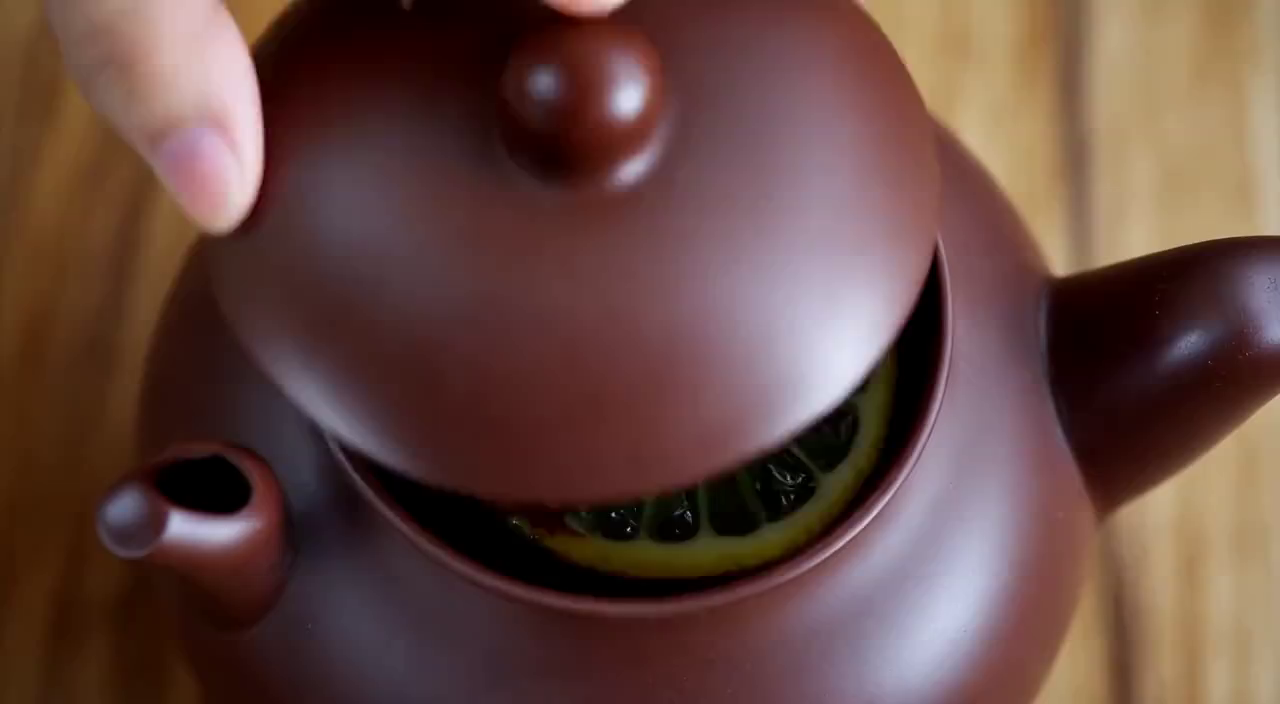}} &
\lpsvideoframe{lpsOurs}{\includegraphics[width=.148\textwidth]{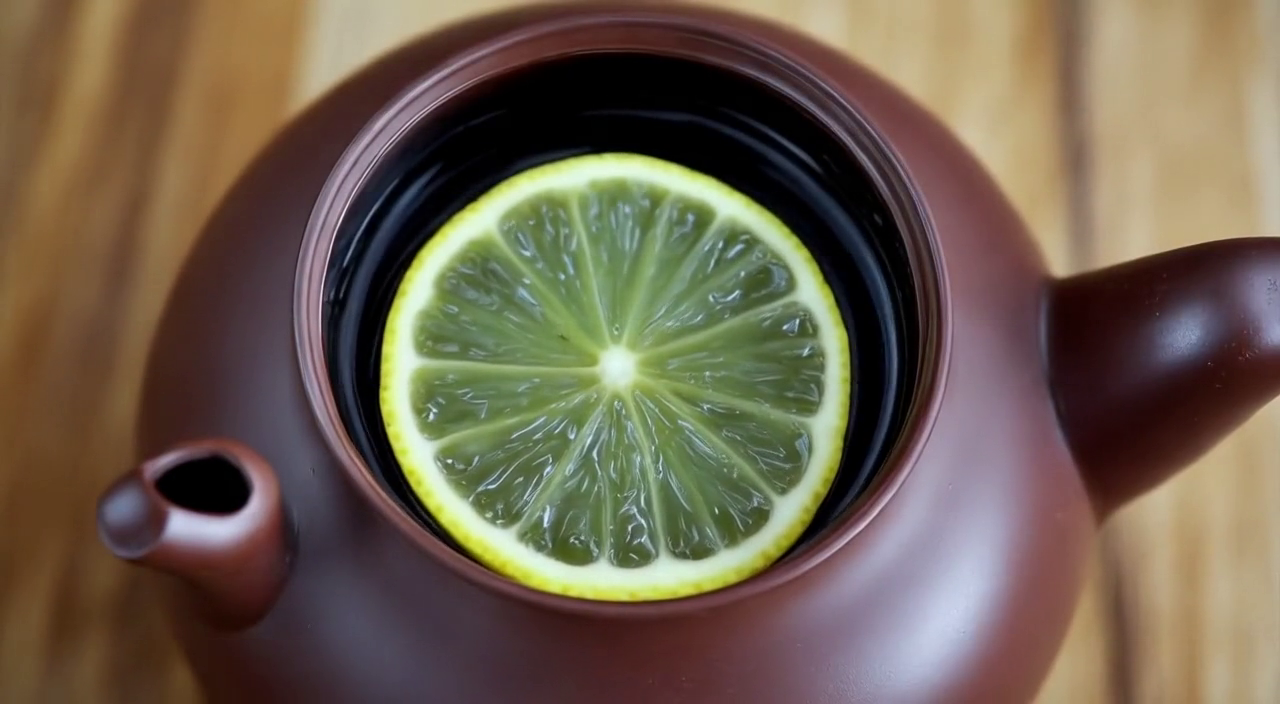}} &
\lpsvideoframe{lpsOurs}{\includegraphics[width=.148\textwidth]{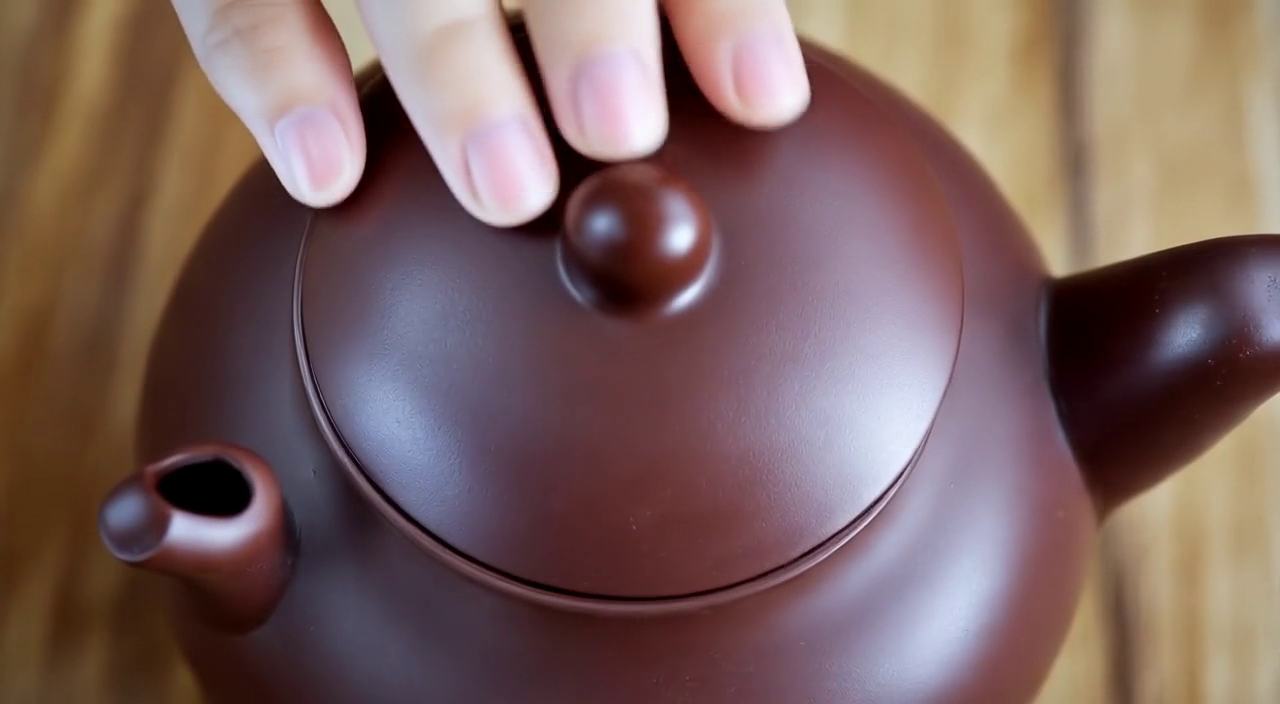}} &
\lpsvideoframe{lpsOurs}{\includegraphics[width=.148\textwidth]{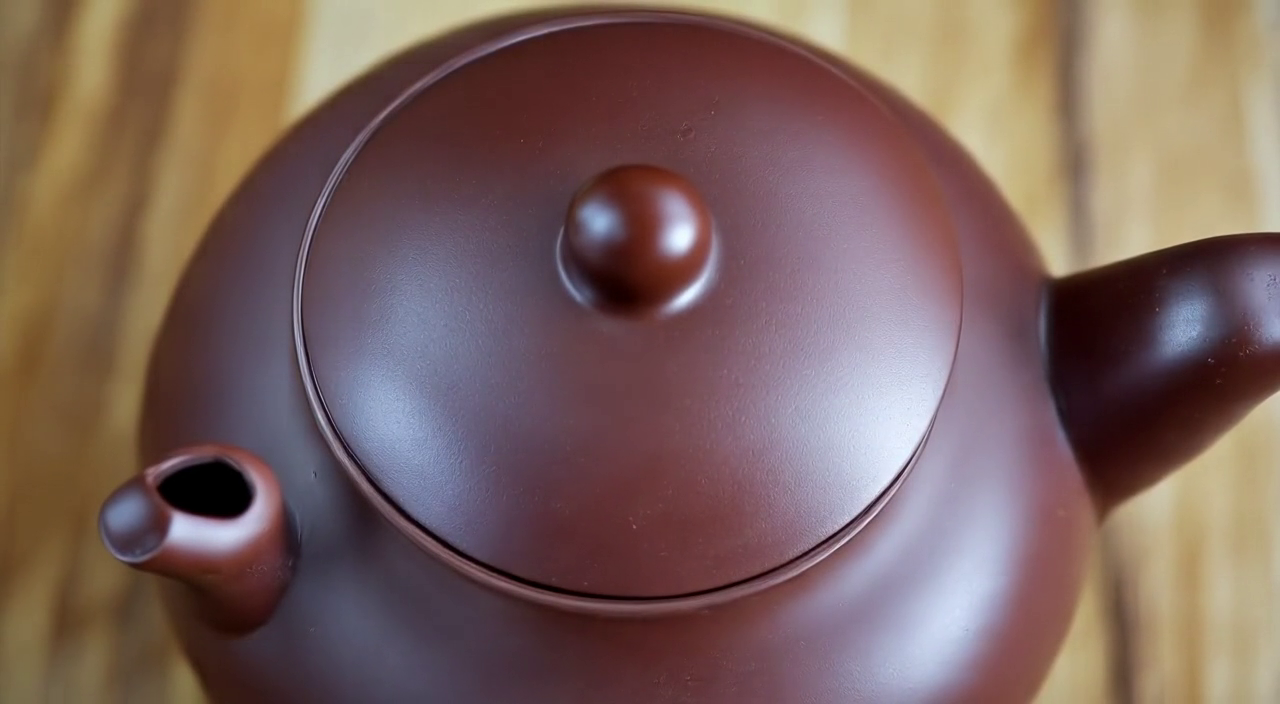}} &
\lpsvideoframe{lpsOurs}{\includegraphics[width=.148\textwidth]{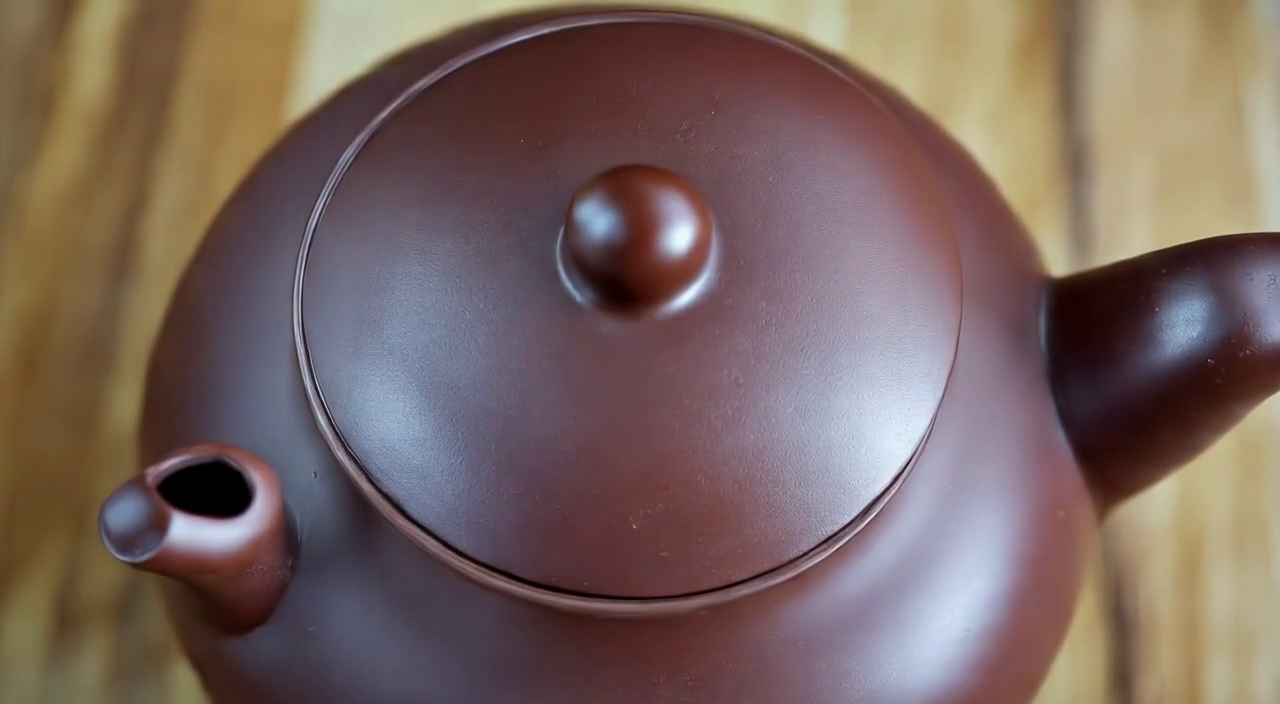}} &
\lpsvideoframe{lpsOurs}{\includegraphics[width=.148\textwidth]{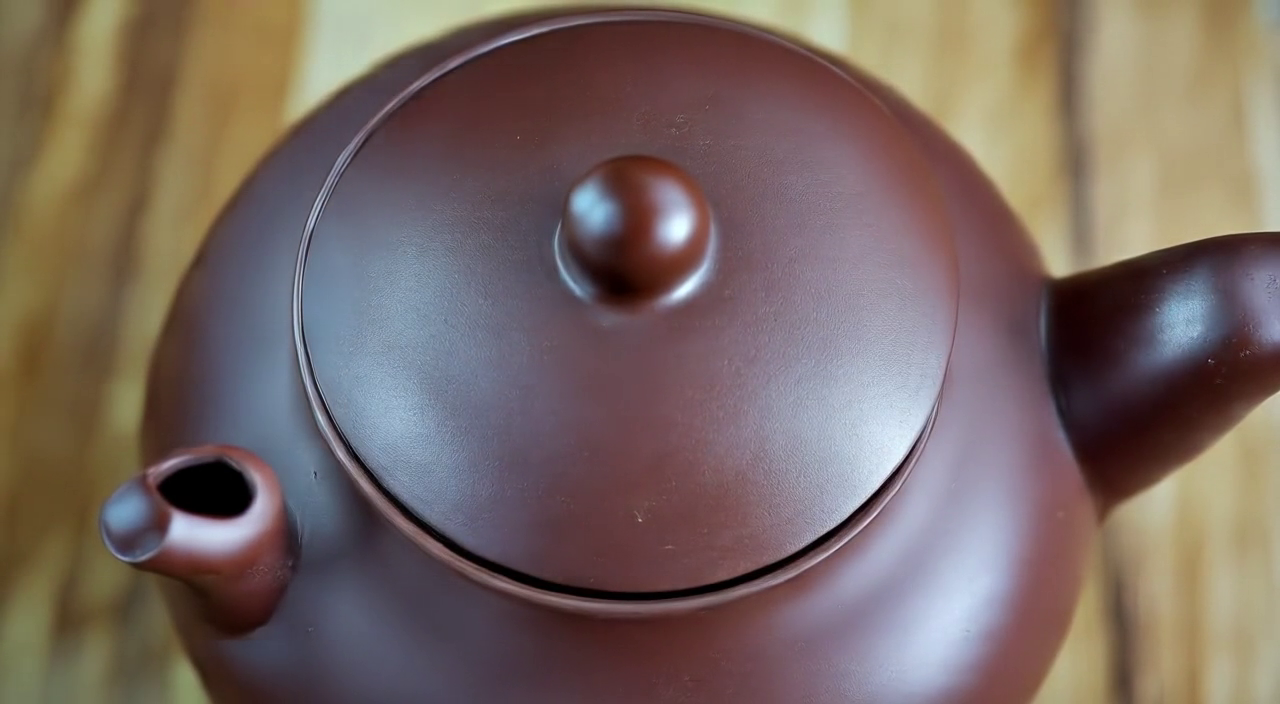}}
\end{tabular*}
\endgroup

\captionsetup{skip=3pt}%
\caption{\textbf{Selected image and video examples with frozen loop programs.} (a,b) Boxes locate enlarged details; \textcolor{lpsOurs}{magenta} frames mark ours. Header numbers are TFLOPs per image; parentheses give DPG scores ($0$--$1$), and \lpsFigCheck{}/\lpsFigCross{} mark GenEval color checks. (c) LongLive~2.0 (5B) frames at matched video times from one RememBench prompt. The baseline's teapot lid deforms and develops white text-like artifacts, while the lid remains nearly unchanged in our frames.}
\label{fig:qualitative}
\end{figure*}

Figure~\ref{fig:qualitative} previews qualitative comparisons across matched prompts and baselines. In panels (a,b), directly reducing standard inference to $10$ steps degrades text spelling and color binding, whereas our $10$-step programs ($\mathrm{ELP}_{\mathrm{speed}}$ on Scale-RAE and $\mathrm{ELP}_{\mathrm{quality}}$ on RAE) preserve both at lower FLOPs than full-step baselines and Loop + LG; panel (c) previews long-video consistency for RQ3. These selected examples do not establish improvement frequency.

\paragraph{Models Trained with Loops.}
\label{sec:exp_pretrained_loops}
Even a backbone pretrained with weight-tied loops under-utilizes its compute when native inference applies uniform depth ($K{=}4$) across flat and high-curvature trajectory regions (Table~\ref{tab:looped_dit_main}). By concentrating steps in high-curvature regions, extrapolating latent updates, and reusing bounded late-loop residuals across neighboring steps, \papershort{} consistently lifts matched-step performance---gaining up to $+4.07$ GenEval and $+3.80$ DPG points at $14$ steps while using less compute, and surpassing the $100$-step reference's GenEval at one-quarter of its FLOPs ($25$ steps), though Full retains a slight DPG edge.

Crucially, these gains do not reduce to timestep warping alone: on $1{,}298$ withheld prompts at matched compute (${\sim}6.8$\,TFLOPs), the complete Speed program still outperforms a schedule-only control ($+1.32$ GenEval, $+1.13$ DPG), confirming that co-designing dynamics and bounded residual reuse contributes beyond scheduling without isolating individual sub-mechanisms (Figure~\ref{fig:teaser}b,c; Supplementary Section~\ref{sec:supp_looped_dit}).

\paragraph{\rqheading{2}{Which Search Design Choices Matter?}}
\label{sec:exp_design_choices}
\label{sec:exp_analysis}

Table~\ref{tab:agent_pareto} ablates search space, diagnostic access, search LLM, target strength, and objectives on Scale-RAE. Compare settings at matched priority (Speed or Quality) within each panel, organized below by the four method components.

\begin{table*}[t]
\centering
\caption{\textbf{\rqhighlight{2}{Program search, choice of search LLM, target strength, and objectives alter the selected Scale-RAE trade-offs.}} Panels vary (a) search space and diagnostic access, (b) search LLM, (c) target strength, and (d) objectives on Scale-RAE. In (a), \emph{Config.} searches a predefined configuration space; \emph{Program} also edits executable code. \cmark/\xmark{} indicate diagnostic access in (a) and active objectives in (d); enabled probes are fixed for Config. and LLM-written for Program. Benchmark scores use the full $1{,}618$-prompt set; \textbf{bold}/\underline{underlined} mark best/second-best within each panel.}
\label{tab:agent_pareto}
\begin{subtable}[t]{0.488\textwidth}
\centering
\caption{\textbf{Search Space and Diagnostic Access.}}
\label{tab:ablation_search_mode}
\label{tab:ablation_diagnostics}
\renewcommand{\arraystretch}{1.18}
\footnotesize
\setlength{\tabcolsep}{6pt}
\begin{tabular}{@{}c c l | c c c@{}}
\toprule
\multirow{2}{*}{\shortstack{\textbf{Search}\\\textbf{type}}} & \multirow{2}{*}{\shortstack{\textbf{Diagnostic}\\\textbf{access}}} & \multirow{2}{*}{\textbf{Priority}} & \textbf{GenEval} & \textbf{DPG} & \multirow{2}{*}{\textbf{TFLOPs $\downarrow$}} \\
 & & & \textbf{(\%) $\uparrow$} & \textbf{(\%) $\uparrow$} & \\
\midrule
Config. & \cmark & Speed & 55.05 & 79.45 & \underline{17.34} \\
Config. & \cmark & Quality & 59.88 & \textbf{80.26} & 29.97 \\
\midrule
Program & \xmark & Speed & 52.41 & 80.05 & \textbf{11.99} \\
Program & \xmark & Quality & 57.63 & 79.68 & 22.08 \\
\midrule
\rowcolor{gray!12}
Program & \cmark & Speed & \underline{60.54} & \underline{80.19} & 19.56 \\
\rowcolor{gray!12}
Program & \cmark & Quality & \textbf{60.98} & \underline{80.19} & 27.13 \\
\bottomrule
\end{tabular}
\end{subtable}%
\hfill
\begin{subtable}[t]{0.488\textwidth}
\centering
\caption{\textbf{How to Search: Choice of Search LLM.}}
\label{tab:ablation_agent_target}
\renewcommand{\arraystretch}{1.18}
\footnotesize
\setlength{\tabcolsep}{3pt}
\begin{tabular*}{\linewidth}{@{\extracolsep{\fill}}l | c c | c c@{}}
\toprule
\multirow{2}{*}{\textbf{Priority}} & \textbf{GenEval} & \textbf{DPG} & \multirow{2}{*}{\textbf{TFLOPs $\downarrow$}} & \textbf{Speedup} \\
 & \textbf{(\%) $\uparrow$} & \textbf{(\%) $\uparrow$} & & \textbf{(Base / w/ LG)} \\
\midrule
\addlinespace[5.3pt]
\multicolumn{5}{@{}l}{\texttt{Gemini-4-Argon}} \\
Speed & 57.30 & 79.64 & \underline{25.53} & \underline{$1.24\times$} / \underline{$3.43\times$} \\
Quality & 58.48 & \underline{80.18} & 41.85 & $0.75\times$ / $2.09\times$ \\
\midrule
\addlinespace[5.3pt]
\multicolumn{5}{@{}l}{\texttt{GPT-6-Astra}} \\
Speed & \underline{60.54} & \textbf{80.19} & \textbf{19.56} & $\boldsymbol{1.61\times}$ / $\boldsymbol{4.48\times}$ \\
Quality & \textbf{60.98} & \textbf{80.19} & 27.13 & $1.16\times$ / $3.23\times$ \\
\bottomrule
\end{tabular*}
\end{subtable}

\par\medskip
\begin{subtable}[t]{0.620\textwidth}
\centering
\caption{\textbf{What to Evaluate: Target Strength.}}
\label{tab:rae_synthetic_target_ablation}
\renewcommand{\arraystretch}{1.10}
\footnotesize
\setlength{\tabcolsep}{2.1pt}
\begin{tabular}{@{}l c l r | c c | c c@{}}
\toprule
\multirow{2}{*}{\textbf{Method}} & \multirow{2}{*}{\shortstack{\textbf{Target}\\\textbf{Strength}}} & \multirow{2}{*}{\textbf{Priority}} & \multirow{2}{*}{\textbf{\#step}} & \textbf{GenEval} & \textbf{DPG} & \multirow{2}{*}{\textbf{TFLOPs $\downarrow$}} & \textbf{Speedup} \\
 &  & & & \textbf{(\%) $\uparrow$} & \textbf{(\%) $\uparrow$} &  & \textbf{(Base / w/ LG)} \\
\midrule
$\mathrm{ELP}$ (Ours) & $0.00\%$ & Quality & 25 & 58.48 & 80.18 & 41.85 & $0.75\times$ / $2.09\times$ \\
$\mathrm{ELP}^{\mathrm{strong}}$ (Ours) & $17.86\%$ & Quality & 20 & \underline{61.03} & \textbf{80.49} & 44.17 & $0.72\times$ / $1.98\times$ \\
\midrule
\rowcolor{gray!12}
 & & \textbf{Speed} & 9 & 58.32 & \underline{80.40} & \textbf{16.09} & $\boldsymbol{1.96\times}$ / $\boldsymbol{5.45\times}$ \\
\rowcolor{gray!12}
\multirow{-2}{*}{$\mathrm{ELP}^{\mathrm{ultra}}$ (Ours)} & \multirow{-2}{*}{$29.28\%$} & \textbf{Quality} & 16 & \textbf{61.63} & 80.22 & \underline{31.23} & \underline{$1.01\times$} / \underline{$2.81\times$} \\
\bottomrule
\end{tabular}
\end{subtable}%
\hfill
\begin{subtable}[t]{0.368\textwidth}
\centering
\caption{\textbf{What to Evaluate: Search Objectives.}}
\label{tab:rae_single_metric_search}
\footnotesize
\setlength{\tabcolsep}{1.2pt}
\renewcommand{\arraystretch}{1.05}
\begin{tabular}{@{}c c c c | c c c@{}}
\toprule
\multicolumn{3}{@{}c}{\textbf{Search Objectives}} & \multicolumn{1}{c}{\multirow{2.2}{*}{\textbf{Priority}}} & \textbf{GenEval} & \textbf{DPG} & \multirow{2.2}{*}{\textbf{\#step $\downarrow$}} \\
\cmidrule(r){1-3}
\textbf{GenEval} & \textbf{DPG} & \textbf{Speed} & \multicolumn{1}{c}{} & \textbf{(\%) $\uparrow$} & \textbf{(\%) $\uparrow$} & \\
\midrule
\cmark & \cmark & \cmark & Speed & 60.54 & \underline{80.19} & \textbf{10} \\
\cmark & \cmark & \cmark & Quality & \underline{60.98} & \underline{80.19} & \underline{14} \\
\midrule
\cmark & \xmark & \xmark & --- & \textbf{63.15} & 80.15 & 20 \\
\xmark & \cmark & \xmark & --- & 56.09 & \textbf{80.61} & 20 \\
\bottomrule
\end{tabular}
\end{subtable}
\end{table*}

\paragraph{What to Search: Configurations vs. Programs.} Fixed parameter menus can tune loop depth and placement, but cannot invent new execution rules such as periodic reference refresh or cross-step correction reuse. Under matched anchors, LLM, and $80$-candidate budget (Table~\ref{tab:ablation_search_mode}), unlocking executable program edits breaks this representational ceiling---lifting Speed GenEval by $+5.49$\,pp over configuration search and expanding the retained development hypervolume frontier (Figure~\ref{fig:teaser}a,e), whereas configuration search trades away compositional fidelity for slightly lower Speed cost or marginal Quality DPG ($+0.07$\,pp). Because cumulative hypervolume is nondecreasing by construction, upward frontier steps reflect retained breakthroughs rather than monotonic per-proposal gains.

\paragraph{What to Observe: Programmable Diagnostics.}
Removing runtime diagnostics while keeping full code-editing freedom reveals why blind program search struggles (Table~\ref{tab:ablation_diagnostics}). Without visibility into internal state drift or guidance stability, scores-only search defaults to aggressive step truncation: it produces a very cheap Speed point ($11.99$\,TFLOPs) that inflates development latency hypervolume ($1.192$ vs.\ $0.946$; Table~\ref{tab:supp_agent_pareto_balanced}), yet fails to discover cross-step guidance reuse and trails even configuration search on compositional fidelity (losing up to $8.13$\,pp GenEval against diagnostic-guided search). Runtime probes steer edits toward safe reuse rules instead of destructive shortcuts (Supplementary Section~\ref{sec:supp_agent_episode}), though these single runs do not isolate fixed versus LLM-written probes or search variance.

\paragraph{How to Search: Choice of Search LLM.}
Because program discovery relies on turning noisy traces into valid code hypotheses, the search frontier is sensitive to the reasoning and synthesis strength of the search LLM. Under identical targets and an $80$-candidate budget (Table~\ref{tab:ablation_agent_target}), \texttt{GPT-6-Astra} consistently finds stronger accuracy--compute trade-offs than \texttt{Gemini-4-Argon} (up to $+3.24$\,pp GenEval at substantially lower compute), illustrating LLM sensitivity while noting that establishing a universal ranking requires repeated matched-budget runs.

\paragraph{What to Evaluate: Target Strength.}
Target ambition acts as a behavioral prior on search: conservative measured targets (True) keep \texttt{Gemini-4-Argon} within familiar multi-step schedules, whereas aspirational targets (Strong and Ultra, Table~\ref{tab:rae_synthetic_target_ablation}) push search toward structural shortcuts---such as strided guidance with velocity-delta reuse under Ultra---that simultaneously lift GenEval and cut inference compute by up to $1.96\times$ over Baseline. Yet target pressure has diminishing and domain-dependent returns: held-out GenEval slightly reverses the Strong/Ultra ordering ($59.53$ vs.\ $59.29$), and tighter discrete text reasoning favors unshifted targets (Supplementary Sections~\ref{sec:supp_rae_synthetic_targets}--\ref{sec:supp_llm_pressure}). Comparing Pareto retention against scalarized selection remains future work.

\paragraph{What to Evaluate: Search Objectives.}
Search objectives directly govern which region of the quality--compute trade-off is explored (Table~\ref{tab:rae_single_metric_search}). Dropping the speed objective lets single-metric searches invest heavier step budgets ($20$ vs.\ $10$--$14$ steps) to specialize on their target metric---pushing GenEval to $63.15$ under GenEval-only or DPG to $80.61$ under DPG-only---but single-metric specialization can erode unmonitored capabilities (DPG-only drops GenEval by nearly $5$\,pp below joint Quality). Because step budgets and selection rules also differ across runs (GenEval-only selects among three held-out replays, so its held-out score is not independent confirmation), these comparisons illustrate objective steering rather than an isolated causal ablation.

\paragraph{Discovered Loop Programs.}
Table~\ref{tab:searched_solutions} previews how \papershort{} adapts structure, dynamics, and control across modalities, contrasting image generation (Scale-RAE) with autoregressive long-video generation (LongLive~1.0, evaluated in \rqtag{3} below; Table~\ref{tab:longlive10_loop_variants}). Across all evaluated image, video, and language backbones, discovered programs converge on shared principles: concentrating loops inside middle-to-post-center semantic blocks during high-noise or high-curvature stages, matching update dynamics to backbone geometry, and decoupling transient loop passes from persistent KV memory or cached conditioning (Supplementary Sections~\ref{sec:supp_programs} and~\ref{sec:supp_human_alignment}).

\begin{table}[t]
\centering
\caption{\textbf{\rqhighlight{2}{\papershort{} discovers different loop and reuse rules across backbones.}} Structure, dynamics, and control summarize two frozen programs; LongLive shows the additional sweep's CLIP-selected configuration. Block and denoising-step indices are zero-based.}
\label{tab:searched_solutions}
\footnotesize
\setlength{\tabcolsep}{1.5pt}
\renewcommand{\arraystretch}{1.18}
\begin{tabular*}{\columnwidth}{@{\extracolsep{\fill}}>{\raggedright\arraybackslash}p{0.22\columnwidth} | >{\raggedright\arraybackslash}p{0.23\columnwidth} >{\raggedright\arraybackslash}p{0.24\columnwidth} >{\raggedright\arraybackslash}p{0.26\columnwidth}@{}}
\toprule
\textbf{Backbone} & \textbf{Structure} & \textbf{Dynamics} & \textbf{Control} \\
\midrule
\mbox{Scale-RAE}~\cite{tong2026scaling} & Layerwise repetition & Damped Euler + guidance & Periodic reference refresh and reuse \\
\midrule
LongLive\newline 1.0 1.3B~\cite{yang2026longlive} & Blocks $12$--$18$; $K{=}4$ & Windowwise Euler ($1/K{=}0.25$) & \textbf{Loop only at steps $\mathbf{0,1}$}; final-pass KV commit \\
\bottomrule
\end{tabular*}
\end{table}

\paragraph{\rqheading{3}{Can \papershort{} Improve Video Generation?}}
\label{sec:exp_world_models}

\paragraph{Long-Video Consistency.}
Autoregressive video generators suffer compounding exposure bias when noisy early estimates enter the causal KV cache, causing scenes to deform after camera departure and revisit (Figure~\ref{fig:qualitative}c). Repeating middle causal blocks during high-noise chunk steps while deferring KV-cache commits to the final pass stabilizes scene geometry before state enters autoregressive memory. Across $100$ RememBench scenes (Table~\ref{tab:world_model}, Figure~\ref{fig:teaser}d), frozen loop programs consistently improve mean and median CLIP, LPIPS, and TempSSIM across LongLive~2.0, LongLive~1.0, and Self-Forcing, cutting mean LPIPS by $61.3\%$ on LongLive~2.0. Crucially, gains transfer cleanly to the $80$ unseen held-out scenes across all three backbones (cutting held-out LPIPS by $59.6\%$ on LongLive~2.0, $40.2\%$ on LongLive~1.0, and $25.1\%$ on an independent $20/80$ Self-Forcing evaluation; Supplementary Section~\ref{sec:supp_longlive10_additional}), though without step reduction this consistency costs additional compute ($1.80\times$ and $1.61\times$ baseline FLOPs on LongLive~2.0 and 1.0). Window and depth sweeps further expose metric-specific preferences between semantic alignment (CLIP) and perceptual fidelity (LPIPS) (Supplementary Section~\ref{sec:supp_longlive10_additional}).

\paragraph{Short-Video Quality--Compute Trade-offs.}
On short-video diffusion (Wan2.1-1.3B, $256$-video VBench), \papershort{} uncovers two complementary regimes rather than a single dominant rule: pairing step reduction ($26$ vs.\ $50$ steps) with targeted cross-attention looping yields a $2.37\times$ FLOPs speedup that improves temporal stability at a modest overall VBench drop ($83.94$ vs.\ $84.92$), whereas full-step multi-window looping prioritizes semantic and scene alignment (overall $85.12$) at extra compute. Neither program dominates every VBench dimension (Supplementary Section~\ref{sec:supp_benchmark_results}).

\paragraph{\rqheading{4}{Does the Approach Extend to Text Reasoning?}}
\label{sec:exp_text_reasoning}
\label{sec:exp_other_generation}

Because Qwen3-4B was pretrained without loops, its frozen residual blocks tolerate far narrower multi-pass perturbations than diffusion denoisers. Replacing single-window Euler repetition with dual-window RK3 updates and read-only intermediate passes (Table~\ref{tab:llm_reasoning}) nonetheless lifts aggregate reasoning accuracy over both standard inference ($+1.67$\,pp Micro Acc) and Fixed Loop~\cite{chen2026trainingfreelooped}, driven mainly by harder knowledge-intensive tasks (GPQA-Main and MMLU-STEM) while trailing Fixed Loop on normalized ARC. On Qwen3-4B, gains remain modest and uneven, absent compute measurements preclude an efficiency claim, and these STEM benchmarks do not assess open-ended generation. Mirroring the unlooped-versus-looped contrast in \rqtag{1}, Supplementary Section~\ref{sec:supp_dory} additionally evaluates a pretrained looped language model (Dory1.0-4B) on open-ended reasoning, coding, and instruction following, where bounded geodesic extrapolation and converged last-loop KV commits allow $K{=}2$ programs to surpass native $K{=}4$ at roughly $1.7\times$ FLOPs and wall-clock speedup. Separate searches across settings demonstrate cross-domain applicability rather than zero-shot program transfer. 

\subsection{Search Cost}
\label{sec:search_cost}

The reported $80$-candidate searches take $3.4$--$7.2$\,hours end-to-end for the Qwen3-4B and RAE studies, using one RTX~5090 for candidate evaluation. This cost is incurred once per setting; frozen programs run without search-LLM calls or token overhead. Supplementary Section~\ref{sec:supp_exp_settings} details the per-backbone timings, hardware, and token usage.

\FloatBarrier

\section{Conclusion}

\paper{} formulates training-free looping as offline discovery of executable
inference programs, jointly editing loop structure, state-update dynamics,
and execution controls while keeping pretrained weights fixed and deploying
frozen programs without search-LLM overhead. Across image diffusion, video
generation, and language reasoning---spanning models trained both with and
without loops---discovered programs repeatedly converge on coherent design
principles: placing loops in middle-to-post-center semantic blocks during
high-noise or high-curvature stages, matching state updates to model
geometry, and decoupling transient inner passes from persistent KV memory
or cached conditioning (Supplementary Section~\ref{sec:supp_human_alignment}).
Important boundaries remain: gains depend on the task and operating point,
no single program dominates every metric, long-video consistency incurs
extra compute, and repeated matched-budget searches are needed to quantify
search variance. More broadly, these results show that frozen foundation
models can be adapted as programmable test-time compute graphs discovered
offline rather than tuned through fixed hand-crafted schedules.

\section*{LLM Usage Statement}
We used OpenAI Codex to edit and refine author-produced text, experimental designs and analyses, figures and visual materials, data-related materials, and research ideas, claims, and conclusions. For related work and citations, and for code and implementation, we used tool-generated work through one-shot or interactive assistance; the authors reviewed, edited, or approved that work. We did not provide the full manuscript or major parts of it to an LLM provider to obtain reviewer-style feedback. The authors take full responsibility for all submitted content, regardless of any LLM or AI-tool use.

{
    \small
    \bibliographystyle{ieeenat_fullname}
    \bibliography{main}
}

\clearpage
\setcounter{page}{1}
\onecolumn
\raggedbottom
\suppressfloats[t]
\begin{nolinenumbers}
\begingroup
\centering
\Large
\textbf{\thetitle}\par
\vspace{0.5em}
Supplementary Material\par
\vspace{1.0em}
\endgroup
\end{nolinenumbers}
\appendix
\setcounter{table}{0}
\setcounter{figure}{0}
\renewcommand{\thetable}{S\arabic{table}}
\renewcommand{\thefigure}{S\arabic{figure}}
\renewcommand{\theHtable}{supp.\arabic{table}}
\renewcommand{\theHfigure}{supp.\arabic{figure}}
\addtocontents{toc}{\protect\setcounter{tocdepth}{2}}
\begingroup
\setcounter{tocdepth}{2}
\makeatletter
\renewcommand*{\l@paragraph}{\@dottedtocline{4}{3.8em}{0em}}
\makeatother
\renewcommand{\contentsname}{Supplementary Contents}
\hypersetup{linktoc=all}
\pdfbookmark[0]{Supplementary Contents}{appendix.contents}
\tableofcontents
\endgroup
\input{5_supp}

\end{document}